\documentclass[letterpaper]{article}
\usepackage{aaai20}
\usepackage{times}
\usepackage{helvet}
\usepackage{courier}
\usepackage[hyphens]{url}
\usepackage{graphicx}
\usepackage{graphicx}
\usepackage{dsfont}
\usepackage{epstopdf}
\usepackage[subrefformat=parens,labelformat=parens]{subcaption}
\usepackage{algorithm,algorithmic}
\usepackage{amsmath}
\usepackage{amssymb}
\usepackage{amsfonts}
\usepackage{xcolor}
\usepackage{appendix}

\newcommand{\prob}[1]{\mathbb{P}\left(#1 \right)}
\newcommand{\cov}[1]{\mbox{Cov}\left(#1 \right)}

\newcommand{\expe}[1]{\mathbb{E}\left[#1 \right]}
\newcommand{\inprod}[1]{\langle #1 \rangle}
\newcommand{\norm}[1]{\lVert #1 \rVert}
\newcommand{\lnorm}[1]{\left\lVert #1\right \rVert}

\newcommand{\lcb}{\mbox{LCB}}
\newcommand{\argmax}{\mbox{argmax}}
\newcommand{\xg}{X^{\mathsf{CE}}}
\newcommand{\xl}{X^{\mathsf{LCB}}}
\newcommand{\xS}{X^{\mathsf{S}}}
\newcommand{\xs}{X_0}
\newcommand{\tr}[1]{\mbox{trace}\left( #1\right)}

\newcommand{\xse}{X^{\mathsf{SE}}}

\newcommand{\sprob}{\delta}
\newcommand{\NT}{\mathcal{N}_T}

\newtheorem{theore}{Theorem}
\newtheorem{corollar}{Corollary}
\newtheorem{lemm}{Lemma}
\newtheorem{definitio}{Definition}
\newtheorem{assumptio}{Assumption}

\usepackage{pdfpages}

\title{Safe Linear Stochastic Bandits}

\author{Kia Khezeli  \and  Eilyan Bitar\\
School of Electrical and Computer Engineering, Cornell University, Ithaca, NY, USA\\
{\texttt{\{kk839,eyb5\}@cornell.edu} }\\
}

\begin{document}
\maketitle
\begin{abstract}
We introduce the safe linear stochastic bandit framework---a generalization of linear stochastic bandits---where, in each stage, the learner is required to select an arm with an expected reward that is no less than a predetermined (safe) threshold with high probability. We assume that the learner initially has knowledge of an arm that is known to be safe, but not necessarily optimal. Leveraging on this assumption, we introduce a learning algorithm that systematically combines known safe  arms with exploratory arms to safely expand the set of safe arms over time, while facilitating safe greedy exploitation in subsequent stages. In addition to ensuring the satisfaction of the safety constraint at every stage of play, the proposed algorithm is shown to exhibit an expected regret that is no more than $O(\sqrt{T}\log (T) )$ after $T$ stages of play.
\end{abstract}

\section{Introduction}
We investigate the role of safety in constraining the design of learning algorithms  within the classical framework of linear stochastic bandits \cite{dani2008stochastic,rusmevichientong2010linearly,abbasi2011improved}. Specifically, we introduce a family of \emph{safe linear stochastic bandit problems} where---in addition to the typical goal of designing learning algorithms that minimize regret---we impose a constraint requiring that an algorithm's stagewise expected reward remains above a predetermined safety threshold with high probability at every stage of play. In the proposed framework, we assume that a ``safe'' baseline arm is initially known, and consider a class of safety thresholds that are defined as fixed cutbacks on the  expected reward of the known baseline arm.  Accordingly, an  algorithm that is deemed to be safe cannot induce stagewise rewards that dip below the baseline reward by more than a fixed amount. Critically, the assumption of a known baseline arm---and the limited capacity for exploration implied by the class of safety thresholds considered---can be leveraged on to initially guide the exploration of   allowable arms by playing combinations of the baseline arm and  exploratory arms in a manner that expands the set of safe arms over time, while simultaneously preserving safety at every stage of play.

There are a variety of real-world applications that might benefit from the  design of stagewise-safe  online learning algorithms  \cite{khezeli2017risk,libubblerank,sui2015safe}. Most prominently, clinical trials have long been used as a motivating application for the multi-armed bandit \cite{berry1985optimal} and linear bandit \cite{dani2008stochastic} frameworks. However, as pointed out by \cite{villar2015multi}: 
``Despite this apparent near-perfect fit between a real-world problem and a mathematical theory, the MABP has yet to be applied to an actual clinical trial.''   One could argue that the ability to provide a learning algorithm  that is guaranteed to be  stagewise safe has the potential to facilitate the utilization of bandit models and algorithms in clinical trials.  More concretely, consider the possibility of using the linear bandit framework to model the problem of optimizing a combination of $d$ candidate treatments for  a specific health issue. In this context, an ``arm'' represents a mixture of  treatments, the ``unknown reward vector'' encodes the effectiveness of each treatment, and the ``reward'' represents a patient's response to a chosen mixture of treatments. In terms of the safety threshold, it is natural to select  the ``baseline arm'' to be the (possibly suboptimal) combination of treatments possessing the largest reward known to date. As it is clearly unethical to prescribe a treatment that may  degrade a patient's health, the stagewise safety constraint studied in this paper can be interpreted as a requirement that a patient's response  to a chosen treatment must be arbitrarily close to that of the baseline treatment, if not better.

\subsection{Contributions}
In this paper, we propose a new learning algorithm that is tailored to the safe linear bandit framework. The proposed algorithm, which we call the \emph{Safe Exploration and Greedy Exploitation} (SEGE) algorithm, is shown to exhibit near-optimal expected regret, while guaranteeing the satisfaction of the proposed safety constraint at every stage of play.  Initially, the SEGE algorithm performs safe exploration by combining the baseline arm with a random exploratory arm that is constrained by an ``exploration budget'' implied by the stagewise safety constraint. Over time, the proposed algorithm systematically expands the family of safe arms in this manner to include new safe arms with expected rewards that exceed the baseline reward level. Exploitation under the SEGE algorithm is based on the certainty equivalence principle. That is, the algorithm  constructs an ``estimate'' of the unknown reward parameter, and selects an arm that is optimal for the given parameter estimate.  The SEGE algorithm only plays the certainty equivalent (i.e., greedy) arm when it is safe---a condition that is determined according to a lower confidence bound on its expected reward. Moreover, the proposed algorithm balances the trade-off between exploration and exploitation by controlling the rate at which information is accumulated over time, as measured by the growth rate of the minimum eigenvalue of the so-called information matrix.\footnote{We note that a closely related class of  learning algorithms, which explicitly control the rate of information gain in this manner, have been previously studied in the context  of dynamic pricing algorithms for revenue maximization  \cite{den2013simultaneously,keskin2014dynamic}.} More specifically, the SEGE algorithm guarantees that the minimum eigenvalue of the information matrix grows at a rate  ensuring that the expected regret of the algorithm is no greater than  $O( \sqrt{T}\log(T))$ after $T$ stages of play. This regret rate that is near optimal in light of $\Omega(\sqrt{T})$  lower bounds previously established in the  linear stochastic bandit literature  \cite{dani2008stochastic,rusmevichientong2010linearly}. 

\subsection{Related Literature}
There is an extensive literature on linear stochastic bandits. For this setting, several algorithms based on the principle of Optimism in the Face of Uncertainty (OFU) \cite{dani2008stochastic,rusmevichientong2010linearly,abbasi2011improved} or Thompson Sampling \cite{agrawal2013thompson} have been proposed.  Although such algorithms are known to be near-optimal under various measures of regret, they may fail in the safe linear bandit framework, as their (unconstrained) approach to exploration may result in a violation of the stagewise safety constraints considered in this paper.

In the context of multi-armed bandits, there is a related stream of literature that focuses on the design of ``risk-sensitive'' learning algorithms by encoding risk in the performance objectives according to which regret is measured \cite{cassel2018general,david2018pac}. Typical risk measures that have been studied in the multi-armed bandit literature include Mean-Variance \cite{sani2012risk,vakili2016risk}, Value-at-Risk \cite{vakili2015mean}, and Conditional Value-at-Risk \cite{galichet2013exploration}. Although such risk-sensitive algorithms are inclined to exhibit reduced volatility in the cumulative reward that is received over time, they are not constrained in a manner that explicitly limits the stagewise risk of the reward processes that they induce.

Closer to the setting studied in this paper is the conservative bandit framework \cite{wu2016conservative,kazerouni2017conservative}, which incorporates explicit safety constraints on the reward process induced by the learning algorithm. However, in contrast to the stagewise safety constraints considered in this paper, conservative bandits encode their safety requirements in the form of  constraints on the cumulative rewards received by the algorithm.  Along a similar line of research,  \cite{sun2017safety} investigate the design of learning algorithms for risk-constrained contextual bandits that balance a tradeoff between cumulative constraint violation and regret. Given the cumulative nature of the safety constraints considered by the aforementioned algorithms, they cannot be directly applied to the stagewise safe linear bandit problem considered in this paper. In Section \ref{sec:comp},  we provide a simulation-based comparison between the SEGE algorithm and the Conservative Linear Upper Confidence Bound (CLUCB) algorithm  \cite{kazerouni2017conservative} to more clearly illustrate the potential weaknesses and strengths of each approach. 

We close this section by mentioing another closely related body of work in the online learning literature that investigates the design of stagewise-safe algorithms for a more general class of smooth reward  functions \cite{sui2015safe,sui2018stagewise,usmanova2019safe}. Although the proposed algorithms are shown to respect stagewise safety constraints that are similar in spirit to the class of safety constraints considered in this paper, they lack formal upper bounds on their cumulative regret.

\subsection{Organization}
The remainder of the paper is organized as follows.  We introduce  pertinent notation in Section \ref{sec:notation}. In Section \ref{sec:model}, we define the safe linear stochastic bandit problem. In Section \ref{sec:SEGE}, we introduce the Safe Exploration and Greedy Exploitation (SEGE) algorithm. We present our main theoretical findings in Section \ref{sec:results}, and close the paper with a simulation study of the SEGE algorithm in Section \ref{sec:numerical}. All mathematical proofs are presented in the Appendix to the paper.

\section{Notation} \label{sec:notation}
We denote the standard Euclidean norm of a vector $x\in \mathbb{R}^d$ by $\norm{x}$ and define its weighted Euclidean norm as $\norm{x}_S = \sqrt{x^\top S x}$ where $S\in \mathbb{R}^{d\times d}$ is a given symmetric positive semidefinite matrix. We denote the inner product of two vectors  $x,y\in \mathbb{R}^d$ by $\inprod{x,y} = x^\top y$. For a square matrix $A\in \mathbb{R}^{d\times d}$, we denote its minimum and maximum eigenvalues by $\lambda_{\min}(A)$ and $\lambda_{\max}(A)$, respectively.

\section{Problem Formulation} \label{sec:model}

In this section,  we  introduce the safe linear stochastic bandit model considered in this paper. Before doing so, we  review the standard model for  linear stochastic bandits on which our formulation is based. 

\subsection{Linear Bandit Model}
Linear stochastic bandits  belong to a class of sequential decision-making problems in which a learner (i.e., decision-maker) seeks to maximize an unknown linear function using noisy observations of its function values that it collects over multiple stages. More precisely, at each  stage $t=1,2,\dots$, the learner is required to select an arm (i.e., action) $X_t$ from a  compact set $\mathcal{X}\subset \mathbb{R}^d$ of allowable arms, which is assumed to be an ellipsoid of the form
\begin{align}
\mathcal{X}=\left\{x\in \mathbb{R}^d\ | \ (x-\bar{x})^\top H^{-1} (x-\bar{x})\leq 1 \right\}, \label{eq:ellipsoid}
\end{align}
where $\bar{x}\in \mathbb{R}^d$ and $H \in \mathbb{R}^{d \times d}$ is a symmetric and positive definite matrix. In response to the particular arm played at each stage $t$, the learner observes a  reward $Y_t$  that is induced by the stochastic linear relationship:
\begin{align}
Y_t=\inprod{X_t,\theta^*} + \eta_t.
\end{align}
Here, the noise process  $\{\eta_t\}_{t=1}^{\infty}$ is assumed be a sequence of independent and zero-mean random variables, and, critically, the reward parameter $\theta^* \in \mathbb{R}^d$ is assumed to be fixed and unknown. This a priori uncertainty in the reward parameter gives rise to the need to  balance the exploration-exploitation trade-off in adaptively guiding the sequence of arms played in order to maximize  the expected reward  accumulated over time.

\subsubsection{Admissible Policies and Regret.} We restrict the learner's decisions to those which are non-anticipating in nature. That is to say, at each stage $t$, the learner is required to select an arm based only on the history of past observations $H_t = (X_1, Y_1, \dots, X_{t-1}, Y_{t-1})$, and  on an external source of randomness  encoded by a random variable $U_t$. The  random process $\{U_t\}_{t=1}^\infty$ is assumed to be independent across time, and independent of the random noise process $\{\eta_t\}_{t=1}^\infty$. Formally, an \emph{admissible policy} is a sequence of functions $\pi = \{\pi_t\}_{t=1}^\infty$, where each function $\pi_t$  maps the information  available to the learner at each stage $t$ to a feasible arm  $ X_t \in \mathcal{X}$ according to $X_t = \pi_t(H_t, U_t)$.

The performance of an admissible policy after $T$ stages of play is  measured according to its \emph{expected regret},\footnote{It is worth noting, that in the context  of linear stochastic bandits, \emph{expected regret} is equivalent to \emph{expected pseudo-regret} due to the additive nature of the noise process \cite{abbasi2011improved}.} which equals the difference between the expected reward accumulated by the optimal arm  and the expected reward accumulated by the given policy after $T$ stages of play. Formally, the expected regret of an admissible policy is defined as
\begin{align}
R_T = \sum_{t=1}^T  \inprod{X^*,\theta^*}  -  \expe{\sum_{t=1}^T \inprod{X_t,\theta^*}}, \label{eq:regret def}
\end{align}
where expectation is taken with respect to the distribution induced by the underling  policy, and  $X^* \in \mathcal{X}$  denotes the  \emph{optimal arm} that maximizes the  expected reward at each stage of play given knowledge of the reward parameter $\theta^*$, i.e.,
\begin{align}
X^* =  \argmax_{x\in\mathcal{X}} \inprod{x,\theta^*}.
\end{align} 
 At a minimum, we seek policies  exhibiting an expected regret that is  sublinear in the number of stages played $T$. Such policies are said to have \emph{no-regret} in the sense that $\lim_{T\rightarrow\infty} \ R_T/T=0$.  To facilitate the design and theoretical analysis of such policies, we adopt a number of technical assumptions, which are standard in the literature on linear stochastic bandits, and are assumed to hold throughout the paper.
  \begin{assumptio} \label{ass:bound_theta} The unknown reward parameter is bounded according to $\|\theta^*\| \leq S$, where $S >0$ is a known constant.
\end{assumptio} 
Assumption \ref{ass:bound_theta} will prove essential to the design of policies that \emph{safely explore} the  parameter space in a manner ensuring that the expected reward stays above a  predetermined (safe) threshold with high probability at each stage of play. We refer the reader to Definition \ref{def:safety} for a formal definition of the particular safety notion considered in this paper.
 
\begin{assumptio} \label{ass:subgaus}
Each element of $\{\eta_t\}_{t=1}^\infty$  is assumed to be $\sigma_\eta$-sub-Gaussian, where $\sigma_\eta \geq 0$ is a fixed constant.  That is,
\begin{align*}
\expe{\exp(\gamma\eta_t)} \leq \exp\left(\gamma^2\sigma_\eta^2/2 \right)
\end{align*}
for all  $\gamma\in \mathbb{R}$ and $t \geq 1$.
\end{assumptio} 
Assumptions  \ref{ass:bound_theta} and \ref{ass:subgaus}, together with the class of admissible policies considered in this paper,  enable the utilization of existing results that provide an explicit characterization of  confidence ellipsoids for the unknown reward parameter based on a $\ell_2$-regularized least-squares estimator \cite{abbasi2011improved}. Such confidence regions play a central role in the design of no-regret algorithms for the linear stochastic bandits  \cite{dani2008stochastic,rusmevichientong2010linearly,abbasi2011improved}.

\subsection{Safe Linear Bandit Model}\label{sec:safe bandit}

In what follows, we introduce the framework of \emph{safe linear stochastic bandits} studied in this paper. Loosely speaking, an admissible policy is said to be \emph{safe} if  the  expected reward $\expe{Y_t\mid X_t} = \inprod{ X_t, \theta^*}$ that it induces at each stage $t$ is guaranteed to stay above a given reward threshold with high probability.\footnote{To simplify the exposition, we will frequently refer to $\expe{Y_t\mid X_t}$---the  expected reward conditioned on the arm $X_t$---as the \emph{expected reward}, unless it is otherwise unclear from the context.} More formally, we have the following definition.

\begin{definitio}[Stagewise Safety Constraint] \label{def:safety}
Let $b\in \mathbb{R}$ and $\sprob\in[0,1]$. An admissible policy $\pi$---or equivalently the arm $X_t$ that it induces---is defined to be \emph{$(\sprob,b)$-safe at stage $t$}  if
\begin{align}
\prob{\inprod{ X_t, \theta^*}\geq b}\geq 1-\sprob, \label{eq:safety cond}
\end{align} 
where the probability is calculated according to the distribution induced by the policy $\pi$.
\end{definitio}
The stagewise safety constraint requires that the expected reward at stage $t$  exceed the \emph{safety threshold} $b \in \mathbb{R}$   with probability no less  than $1- \delta$, where  $\delta \in [0,1]$ encodes the \emph{maximum allowable risk} that the learner is willing to tolerate.

Clearly, without making additional assumptions, it is not possible to design policies that are guaranteed  to be safe according to \eqref{eq:safety cond} given  arbitrary safety specifications. We circumvent this obvious limitation by giving  the learner access to a \emph{baseline arm} with a known lower bound on its expected reward. We formalize this assumption as follows.

\begin{assumptio}[Baseline Arm] \label{ass:baseline}
We assume that the learner  knows a deterministic baseline arm $\xs\in\mathcal{X}$ satisfying  $$\inprod{\xs,\theta^*}\geq b_0,$$
where  $b_0 \in \mathbb{R}$ is  a known lower bound on its expected reward. 
\end{assumptio} 
We note that it is  straightforward to construct a baseline arm satisfying Assumption \ref{ass:baseline} by leveraging on the assumed boundedness of the unknown reward parameter as specified by Assumption \ref{ass:bound_theta}. In particular, any arm $X_0 \in \mathcal{X}$ and its corresponding ``worst-case'' reward given by  $b_0 = \min_{\|\theta\| \leq S}  \ \inprod{X_0, \theta} = -S \|X_0\|$ are guaranteed to satisfy Assumption \ref{ass:baseline}.

With Assumption \ref{ass:baseline} in hand, the learner can leverage on the baseline arm to initially guide its exploration of allowable arms by playing combinations of the baseline arm and carefully designed exploratory arms in a manner that safely expands the set of safe arms over time. Plainly, the ability to safely explore in the vicinity of the baseline arm  is only possible under stagewise safety constraints defined in terms of safety thresholds satisfying $b < b_0$. Under such stagewise safety constraints, the difference in rewards levels $b_0 - b$ can be interpreted as a stagewise ``exploration budget'' of sorts, as it reflects the maximum relative loss in expected reward that the learner is willing to tolerate when playing arms that deviate from the baseline arm. Naturally, the larger the exploration budget, the more aggressively can the learner explore. With the aim of  designing safe learning algorithms that leverage on this simple idea, we will restrict our attention to stagewise safety constraints that are specified in terms of safety thresholds satisfying $b < b_0$.

Before proceeding, we briefly summarize the framework of \emph{safe linear stochastic bandits} considered in this paper. Given a baseline arm satisfying Assumption \ref{ass:baseline}, the learner is initially required to fix a safety threshold  that satisfies $b < b_0$.  At each subsequent stage $t =1,2, \dots$, the learner must select a risk level $\delta_t \in [0,1]$ and a corresponding arm $X_t \in \mathcal{X}$ that is $(\delta_t, b)$-safe. 
The learner aims to design an admissible policy that minimizes its expected regret, while simultaneously ensuring that all arms played satisfy the stagewise safety constraints. In the following section, we propose a policy that is guaranteed to both exhibit no-regret and satisfy the safety constraint at every stage of play.

\subsubsection{Relationship to Conservative  Bandits.} We  briefly discuss the relationship between the safety constraints considered in this paper and the conservative  bandit framework  orginally studied by \cite{wu2016conservative} in the context of multi-armed bandits, and subsequently extended to the setting of linear bandits  by \cite{kazerouni2017conservative}. In contrast to the stagewise safety constraints considered in this paper, conservative bandits encode their safety requirements in the form of  constraints on the cumulative expected rewards received by a policy.  Specifically, given a baseline arm satisfying Assumption \ref{ass:baseline}, an admissible policy is said to respect the safety constraint defined in  \cite{kazerouni2017conservative} if
\begin{align}
\prob{\sum_{k=1}^t \inprod{X_k,\theta^*}\geq (1-\alpha) \sum_{k=1}^t b_0,\ \forall\ t\geq 1}\geq 1-\delta,  \label{eq:conservative const}
\end{align} 
where $\delta \in [0,1]$ and $\alpha \in (0,1)$. Here,  the parameter $\alpha$ encodes the maximum fraction of the cumulative baseline rewards that the learner is willing to forgo over time.  In this context, smaller values of $\alpha$ imply greater levels of conservatism (safety). It is straightforward to show that conservative performance constraints of the form \eqref{eq:conservative const} are a special case of the class of stagewise safety constraints considered in Definition \ref{def:safety}. In particular, if we set the safety threshold  according to  $b = (1 - \alpha) b_0$, and let $\{\delta_t\}_{t=1}^\infty$ be any summable sequence of risk levels satisfying $\sum_{t=1}^\infty \delta_t \leq \delta$, then any admissible policy that is $(\delta_t, b)$-safe for each stage $t \geq 1$ also satisfies  the conservative performance constraint \eqref{eq:conservative const}.

\section{A Safe Linear Bandit Algorithm} \label{sec:SEGE}
In this section, we propose a new algorithm, which we call the \emph{Safe Exploration and Greedy Exploitation} (SEGE) algorithm, that is guaranteed to be safe in every stage of play, while exhibiting a near-optimal expected regret.  Before proceeding with a detailed description of the proposed algorithm, we briefly summarize the basic elements underpinning its design. Initially, the SEGE algorithm performs safe exploration by playing convex combinations of the baseline arm and random exploratory arms in a manner that satisfies Definition \ref{def:safety}. Through this process of exploration, the SEGE algorithm is able to  expand the family of safe arms to incorporate new  arms  that are guaranteed to outperform the baseline arm with high probability. Among all safe arms available to the algorithm at any given stage of play, the  arm with the largest lower confidence bound on its expected reward is used as the basis for safe exploration. The SEGE algorithm performs exploitation by playing the certainty equivalent (greedy) arm based on a $\ell_2$-regularized  least-squares estimate of the unknown reward parameter. The SEGE algorithm only plays the greedy arm when it is safe, i.e.,  when a lower confidence bound on its expected reward exceeds the given safety threshold. Critically, the proposed algorithm balances the trade-off between exploration and exploitation by explicitly controlling the growth rate of the so-called information matrix (cf. Eq.  \eqref{eq:infomat}) in a manner that ensures that the expected regret of the SEGE algorithm is no  greater than  $O(\sqrt{T}\log(T))$ after $T$ stages of play. The pseudocode for the SEGE algorithm is presented in Algorithm \ref{alg:safe alg}. 

In the following section, we introduce a regularized least-squares estimator that will serve as the foundation for the proposed learning algorithm.

\subsection{Regularized Least Squares Estimator} \label{sub:lse}
The $\ell_2$-regularized least-squares estimate of the unknown reward parameter $\theta^*$ based on the information available to the algorithm  up until and including stage $t$  is defined as
\begin{align*}
\widehat{\theta}_t = \underset{\theta\in \mathbb{R}^d}{\mbox{argmin}}\ \left\{\sum_{k=1}^{t} (Y_k-\inprod{X_k,\theta})^2 + \lambda \norm{\theta}^2\right\}.
\end{align*}
Here, $\lambda>0$ denotes a user-specified regularization parameter.
It is straightforward to show that
\begin{align}
\widehat{\theta}_t = V_t^{-1} \sum_{k=1}^{t} X_k Y_k, \label{eq:estimate}
\end{align}
where 
\begin{align} \label{eq:infomat}
V_t = \lambda I +  \sum_{k=1}^{t} X_k X_k^\top.
\end{align}
Throughout the paper, we will frequently refer to the matrix $V_t$ as the \emph{information matrix} at each stage $t$.

The following result taken from  \cite[Theorem 2]{abbasi2011improved}  provides an ellipsoidal characterization of a confidence region for the unknown reward parameter based on the regularized least-squares estimator  \eqref{eq:estimate}.  It is straightforward to verify that the conditions of \cite[Theorem 2]{abbasi2011improved} are satisfied under the standing assumptions of this paper.

\begin{theore}\label{thm:yadkori}
For any admissible policy and  $\delta\in(0,1)$, it holds  that
\begin{align*}
\prob{\theta^*\in \mathcal{C}_t(\delta),\ \forall t\geq 1}\geq 1-\delta,
\end{align*}
where the confidence set $\mathcal{C}_t(\delta)$ is defined as
\begin{align}
\mathcal{C}_t(\delta) = \left\{ \theta\in \mathbb{R}^d\ : \ \norm{\widehat{\theta}_t-\theta}_{V_t} \leq r_t(\delta) \right\}. \label{eq:C_t}
\end{align}
Here, $r_t(\delta)$ is defined as
\begin{align}
r_t(\delta) = \sigma_\eta\sqrt{d\log\left(\frac{1+tL^2/\lambda }{\delta}\right)} + \sqrt{\lambda}S, \label{eq:r_t}
\end{align}
where $L=\max_{x\in\mathcal{X}}{\norm{x}}$.
\end{theore}

In the following section, we propose a method for safe exploration using the characterization of the confidence ellipsoids introduced in Theorem \ref{thm:yadkori}.

\subsection{Safe Exploration}
\begin{figure}
\centering
\includegraphics[width=0.725\columnwidth]{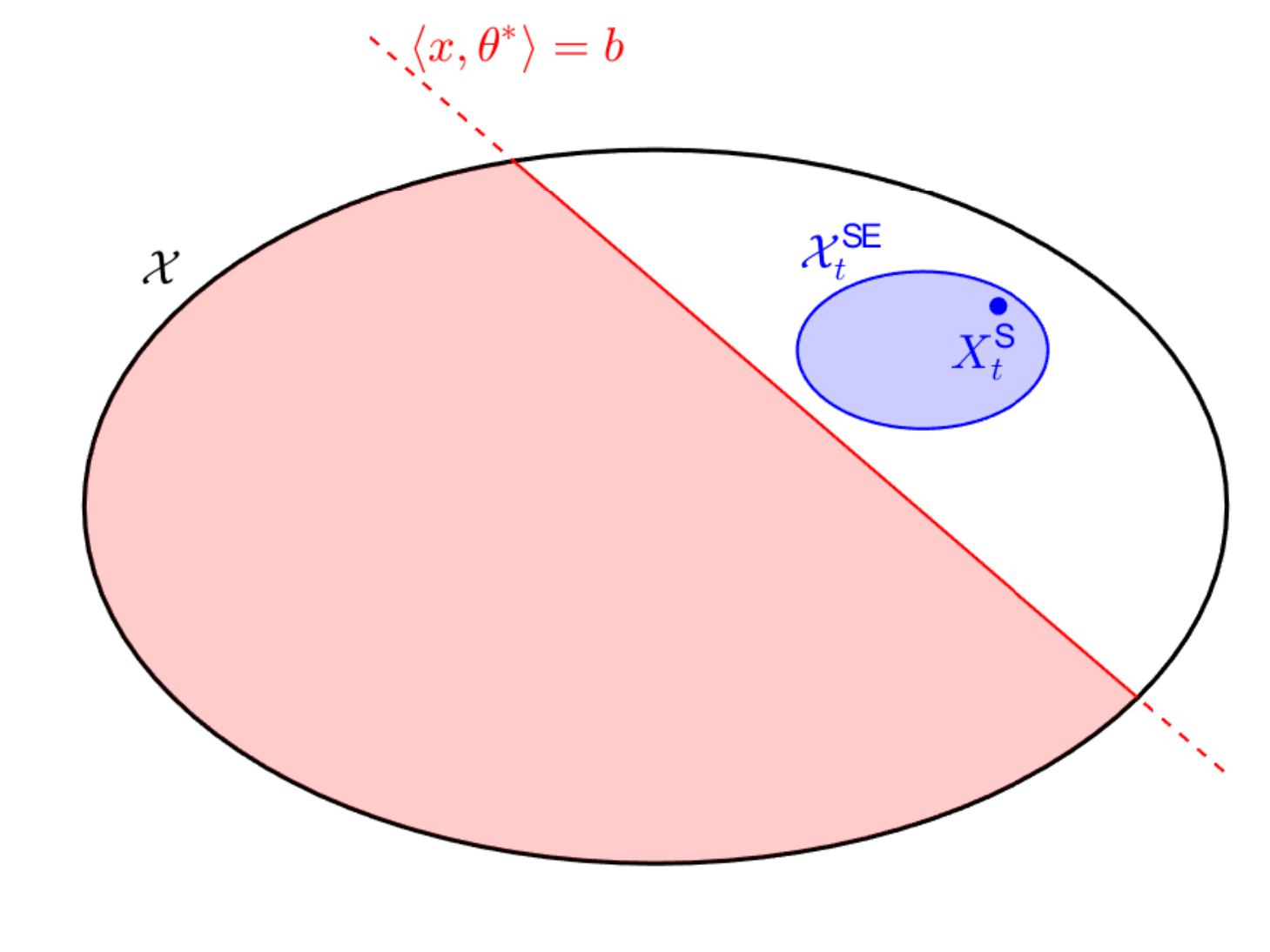}
\caption{The figure illustrates the effect of the safety constraint on the learner's decision making ability. The shaded blue ellipse $\mathcal{X}^{\mathsf{SE}}_t$ depicts the set of all safe exploration arms constructed using the safe arm $\xS_t$ under the SEGE algorithm, i.e., $\mathcal{X}^{\mathsf{SE}}_t = \{(1-\rho)\xS_t+\rho x\mid \rho\in(0,\bar{\rho}), x\in\partial \mathcal{X} \}$. The red shaded area depicts the set of unsafe arms. The black ellipse (and its interior) depicts the set of feasible arms. } \label{fig:safe arm}
\end{figure}

We now describe the approach to ``safe exploration'' that is employed by the proposed algorithm. At each stage $t\geq 1$, given a risk level $\sprob_t$, the SEGE algorithm constructs a safe exploration arm  ($\xse_t$) as a convex combination of a $(\sprob_t,b_0)$-safe arm ($\xS_t$) and a random exploratory arm ($U_t$), i.e.,
\begin{align}
\xse_t = (1-\rho)\xS_t + \rho U_t. \label{eq:safe exp arm}
\end{align}
Qualitatively, the user-specified parameter $\rho\in(0,1)$ controls the  balance between safety and exploration.  Figure \ref{fig:safe arm} provides a graphical illustration of the set of all safe exploration arms induced by a given safe arm $\xS_t$ according to \eqref{eq:safe exp arm}. 

The random exploratory arm process $\{U_t\}_{t=1}^\infty$ is generated according to
\begin{align}
U_t = \bar{x}+H^{1/2}\zeta_t, \label{eq:U_t}
\end{align}
where the random process $\{\zeta_t\}_{t=1}^\infty$ is assumed to be a sequence of independent, zero-mean, and symmetric random vectors. For each element of the sequence, we require that $\norm{\zeta_t}=1$ almost surely and $\sigma_\zeta^2=\lambda_{\min} (\cov{\zeta_t })>0$. Additionally, we define $\sigma^2=\lambda_{\min} (\cov{U_t })$. The parameters $\sigma$ and $\rho$ both determine  how aggressively the algorithm can explore the set of allowable arms. However, exploration that is too aggressive may result in a violation of the stagewise safety constraint. In the following Lemma, we establish an upper bound on $\rho$ such that for all choices of $\rho \in (0, \bar{\rho})$, the arm $\xse_t$ is guaranteed to be safe for any $\sigma\geq 0$. 
\begin{lemm}\label{lem:rho}
Let $\rho\in (0,\bar{\rho})$ where $\bar{\rho} >0$ is defined as
\begin{align}
\bar{\rho} = \min\ \left\{1,\frac{b_0-b }{2S\sqrt{\lambda_{\max}(H)}} \right\}. \label{eq:rho bar}
\end{align} 
Then, for every stage $t\geq 1$, the safe exploration arm $\xse_t$ defined in Equation \eqref{eq:safe exp arm} is $(\sprob,b)$-safe for any $\sprob \in [0,1]$.
\end{lemm}

As the SEGE algorithm expands its set of safe arms over time, it attempts to increase the stagewise efficiency with which it safely explores by exploring in the vicinity of the safe arm with the largest lower confidence bound on its expected reward. More specifically, at each stage $t$,  the SEGE algorithm constructs a confidence set $\mathcal{C}_{t-1}(\sprob_t)$ according to Equation \eqref{eq:C_t}. With this confidence set in hand, the proposed algorithm  calculates a  lower confidence bound (LCB) on the expected reward $\lcb_t(x)$ of each arm $x\in\mathcal{X}$ according to
\begin{align*}
\lcb_t(x) = \min_{\theta\in \mathcal{C}_{t-1}(\sprob_t)}\ \inprod{x,\theta}.
\end{align*}
It is straightforward to show that the  lower confidence bound defined above can be simplified to: 
\begin{align*}
\lcb_t(x) = \inprod{x,\widehat{\theta}_{t-1}}- r_t(\sprob_t)\norm{x}_{V_{t-1}^{-1}}.
\end{align*}
We define the LCB arm ($\xl_t$)  to be  the arm with the largest lower confidence bound on its expected reward among all allowable arms. It is given by:
\begin{align}
\xl_t = \argmax_{x\in\mathcal{X}}\ \mbox{LCB}_t(x). \label{eq:lcb arm}
\end{align}
Clearly,  the LCB arm is guaranteed to be $(\sprob_t,b_0)$-safe if   $\lcb_t(\xl_t)\geq b_0$.
 In this case, the SEGE algorithm relies on the LCB arm for safe exploration, as its expected reward is \emph{potentially} superior to the baseline arm's expected reward.\footnote{It is important to note that the condition $\lcb_t(\xl_t)\geq b_0$ does not guarantee superiority of the LCB arm to the baseline arm, as $b_0$ is only assumed to be a lower bound on the baseline arm's expected reward.} 
Putting everything together,  the SEGE algorithm sets the  safe arm ($\xS_t$)  at each stage $t$ according to:
\begin{align}
\xS_t = \begin{cases}
\xl_t,& \text{if} \ \ \mbox{LCB}_t(\xl_t)\geq b_0,\\
\xs,& \text{otherwise}. \label{eq:safe arm t}
\end{cases}
\end{align}

Before closing this section, it is important to note that the LCB arm \eqref{eq:lcb arm} can be calculated in polynomial time by solving  a second-order cone program. This is in stark contrast to the non-convex optimization problem that needs to be solved when computing the UCB arm (i.e., the arm with the largest upper confidence bound on the expected reward)---a problem that has been shown to be NP-hard in general \cite{dani2008stochastic}.

\subsection{Safe Greedy Exploitation}
We now describe the  method for exploitation  employed by the SEGE algorithm. 
 Exploitation  under the SEGE algorithm  relies on the certainty equivalence principle. That is, the algorithm first estimates the unknown reward parameter according to Equation \eqref{eq:estimate}. Then, the algorithm chooses an arm  that is optimal for the given parameter estimate. Given the ellipsoidal structure of the set of allowable arms, the optimal arm $X^*$ can be calculated as
\begin{align}
X^* =\bar{x}+\frac{H\theta^*}{\norm{\theta^*}_H}. \label{eq:x opt}
\end{align}
Similarly, the certainty equivalent (greedy) arm can be calculated as
\begin{align} 
\xg_t = \bar{x}+\dfrac{H\widehat{\theta}_{t-1}}{\norm{\widehat{\theta}_{t-1}}_H}, \label{eq:greedy arm}
\end{align}
where $\widehat{\theta}_{t-1}$ is the regularized least-squares estimate of the unknown reward parameter, as defined in Equation \eqref{eq:estimate}.

It is important to note that the SEGE algorithm only plays the greedy arm  \eqref{eq:greedy arm} when the lower confidence bound on its expected reward is greater than or equal to the safety threshold $b$. This ensures that the greedy arm is only played when it is safe.

\begin{algorithm}[h]
\caption{SEGE  Algorithm}\label{alg:safe alg}
\vspace{1ex}
\begin{algorithmic}[1]
\STATE \textbf{Input:} $\xs$, $b_0$, $\mathcal{X}$, $S>0$, $c>0$, $\lambda>0$, $b<b_0$, $\rho \in (0,\bar{\rho})$, $\sprob_t\in [0,1]\ \forall t\geq 1$ \\
\STATE \textbf{for $t=1,2,3,\ldots$ do}
\\  \emph{\{Parameter Estimation\}}
\STATE \qquad Set $\widehat{\theta}_{t-1 }$ according to Eq. \eqref{eq:estimate}
\STATE \qquad Set $\mathcal{C}_{t-1}(\sprob_t)$ according to Eq. \eqref{eq:C_t}
\\  \emph{\{Safe Greedy Exploitation\}}
\STATE \qquad \textbf{if} $\lcb_t(\xg_t)\geq b $ \textbf{and} $\lambda_{\min}(V_t)\geq c\sqrt{t} $ 
\STATE \qquad \qquad Set $X_t = \xg_t$ according to Eq.  \eqref{eq:greedy arm}
\\  \emph{\{Safe Exploration\}}
\STATE \qquad \textbf{else}
\STATE \qquad \qquad Set $X_t = \xse_t$ according to Eq. \eqref{eq:safe exp arm} 
\STATE \qquad \textbf{end if}
\STATE \qquad Observe $Y_t=\inprod{X_t,\theta^*}+\eta_t$
\STATE \textbf{end for}
\end{algorithmic}
\end{algorithm}

\section{Theoretical Results} \label{sec:results}
We now present our main theoretical results showing that the SEGE algorithm  exhibits near optimal regret for a large class of risk levels (cf. Theorem \ref{thm:regret}), in addition to being safe at every stage of play (cf. Theorem \ref{thm:safety}). As an immediate corollary to Theorem  \ref{thm:regret}, we establish sufficient conditions under which the SEGE algorithm is also guaranteed to satisfy  the conservative bandit constraint  \eqref{eq:conservative const}, while preserving the upper bound on regret  in Theorem  \ref{thm:regret} (cf. Corollary \ref{cor:kazerouni}).

\begin{theore}[Stagewise Safety Guarantee] \label{thm:safety}
The SEGE algorithm is $(\sprob_t,b)$-safe at each stage, i.e.,
\begin{align*}
\prob{\inprod{ X_t, \theta^*}\geq b}\geq 1-\sprob_t
\end{align*}
for all $t \geq 1$.
\end{theore}

The ability to enforce safety in the sequence of arms played is not surprising given the assumption of a known baseline arm that is guaranteed to be safe at the outset. However, given the potential suboptimality of the baseline arm, a na\"ive policy that plays the baseline arm at every stage will likely incur an expected regret that grows linearly with the number of stages played $T$.  
In constrast, we show, in Theorem \ref{thm:regret}, that the SEGE algorithm exhibits an expected regret that  is no greater than $O(\sqrt{T}\log(T))$ after $T$ stages---a regret rate that is near optimal given existing $\Omega(\sqrt{T})$   lower bounds on regret    \cite{dani2008stochastic,rusmevichientong2010linearly}.

\begin{theore}[Upper Bound on Expected Regret]\label{thm:regret}
 Fix $\overline{\sprob} \in (0,1]$ and $K \geq 0$.  Let  $\{\sprob_t\}_{t=1}^\infty$ be any sequence of risk levels satisfying
\begin{align}
\sprob_t\geq \overline{\sprob} e^{-K\sqrt{t}} \label{eq:risk assumption}
\end{align}
for all $t\geq 1$. Then, there exists finite positive constant $C$  such that the expected regret of the SEGE algorithm is upper bounded as
\begin{align}
 R_T \leq C  \sqrt{T}\log(T)  \label{eq:regret bound}
\end{align} 
for all $T\geq 1$.
\end{theore}
In what follows, we provide a high-level sketch of the proof of Theorem \ref{thm:regret}. The complete proof is presented in Appendix \ref{app:regret}. We bound the expected regret incurred during the safe exploration and the greedy exploitation stages separately. First, we show that the stagewise expected regret incurred when playing the greedy arm is proportional to the mean squared parameter estimation error. We then employ Theorem \ref{thm:yadkori} to show that, conditioned on the event $\{\lambda_{\min}(V_t)\geq c\sqrt{t}\}$, the mean squared parameter estimation error at each stage $t$ is no greater than $O(\log(t)/\sqrt{t})$. It follows that the cumulative expected regret incurred during the exploitation stages is no more than $O(\sqrt{T}\log(T))$ after $T$ stages of play.  Now, in order to upper bound the expected regret accumulated during the safe exploration stages, it suffices to upper bound the expected number of safe exploration stages, since the stagewise regret can be upper bounded by a finite constant under any admissible policy.  We show that the expected number of safe exploration stages is no more than $O(\sqrt{T})$ after $T$ stages of play for any sequence of risk levels that does not decay faster than the rate specified in \eqref{eq:risk assumption}.

We close this section with a result  establishing sufficient conditions under which the SEGE algorithm is guaranteed to satisfy  the conservative performance constraint  \eqref{eq:conservative const}, in addition to being stagewise safe, while satisfying an upper bound on its expected regret that matches that of the CLUCB algorithm \cite{kazerouni2017conservative}[Theorem 5].  Corollary \ref{cor:kazerouni} is stated without proof, as it is an immediate consequence of Theorems \ref{thm:safety} and \ref{thm:regret}.

\begin{corollar}[Conservative Performance Guarantee]\label{cor:kazerouni}
Let $\delta\in (0,1)$. Assume, in addition to the  standing assumptions of Theorem \ref{thm:regret}, that $\{\sprob_t\}_{t=1}^\infty$ is a summable sequence satisfying $\sum_{t=1}^\infty \sprob_t \leq \delta$. Then, the SEGE algorithm satisfies the conservative performance constraint \eqref{eq:conservative const}, and exhibits an expected regret that is upper bounded by $O( \sqrt{T}\log(T))$  for all $T \geq 1$.   
\end{corollar}

\begin{figure*}[t]
    \centering
    
    \begin{subfigure}[t]{0.32\textwidth}
        \includegraphics[width=1.\textwidth]{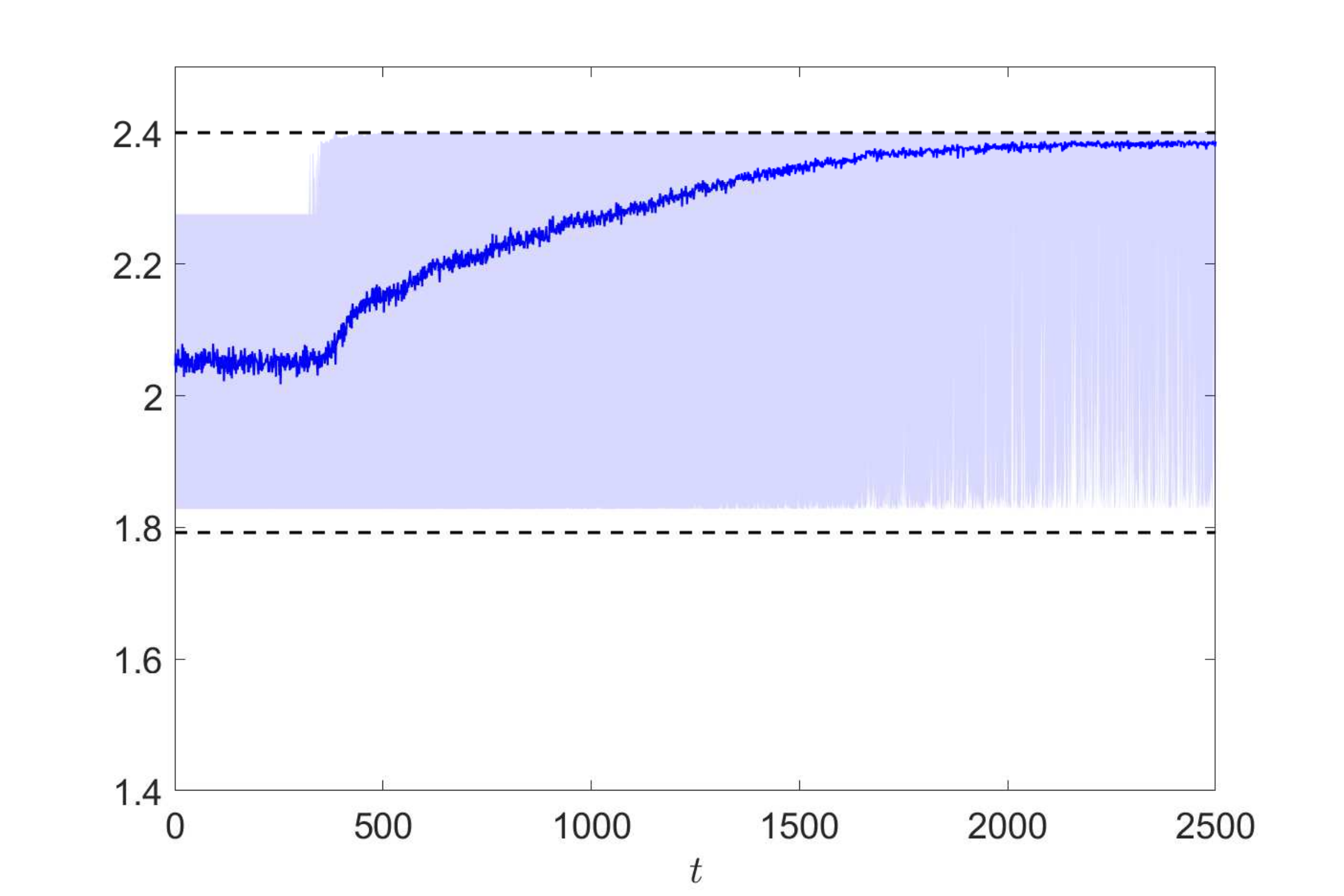}
        \caption{Stagewise expected reward under the SEGE algorithm. }
        \label{fig:sege}
    \end{subfigure}
    ~ 
    \begin{subfigure}[t]{0.32\textwidth}
        \includegraphics[width=1.\textwidth]{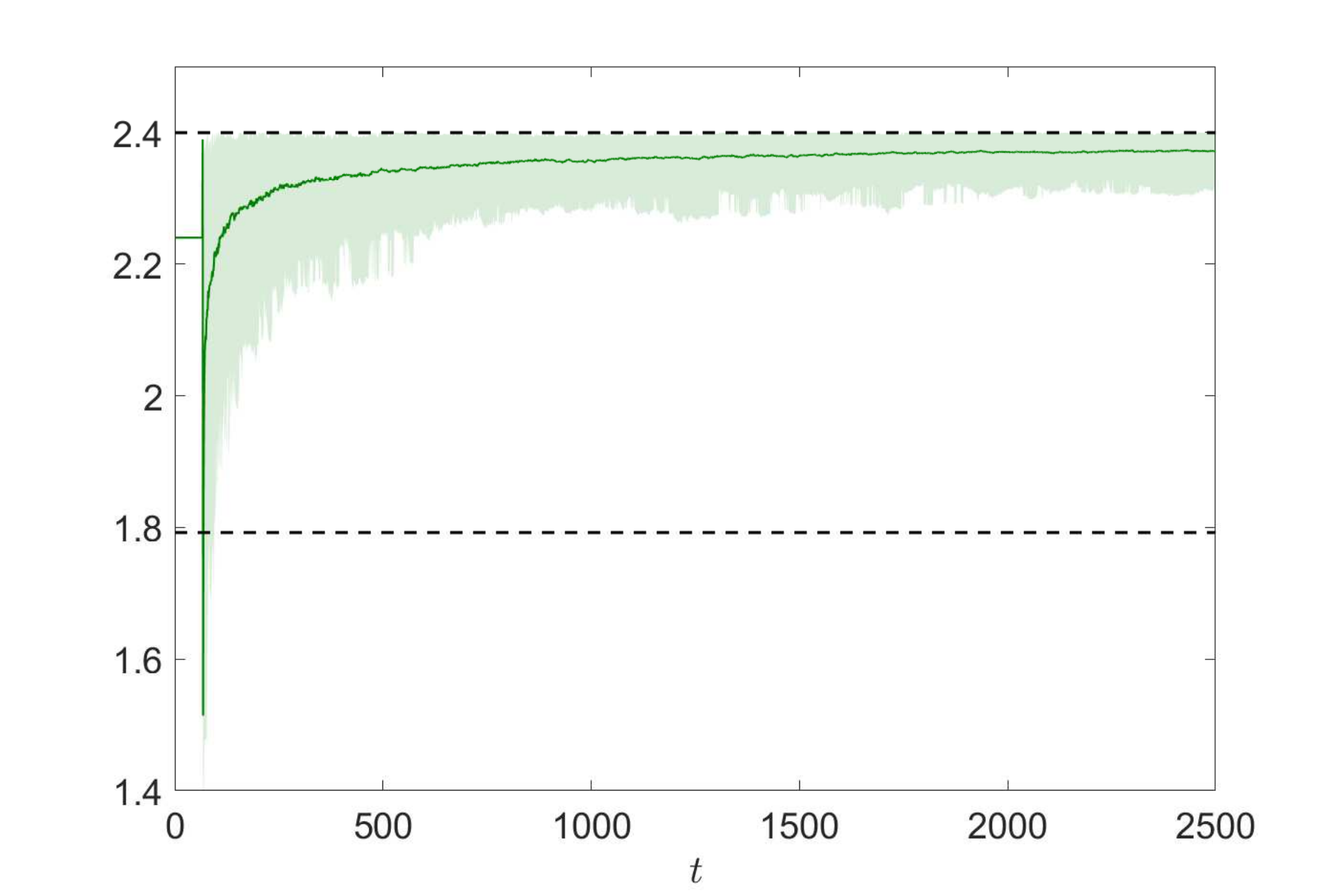}
        \caption{Stagewise expected reward  under the CLUCB algorithm. }
        \label{fig:clucb}
    \end{subfigure}
    ~ 
    \begin{subfigure}[t]{0.32\textwidth}
        \includegraphics[width=1.\textwidth]{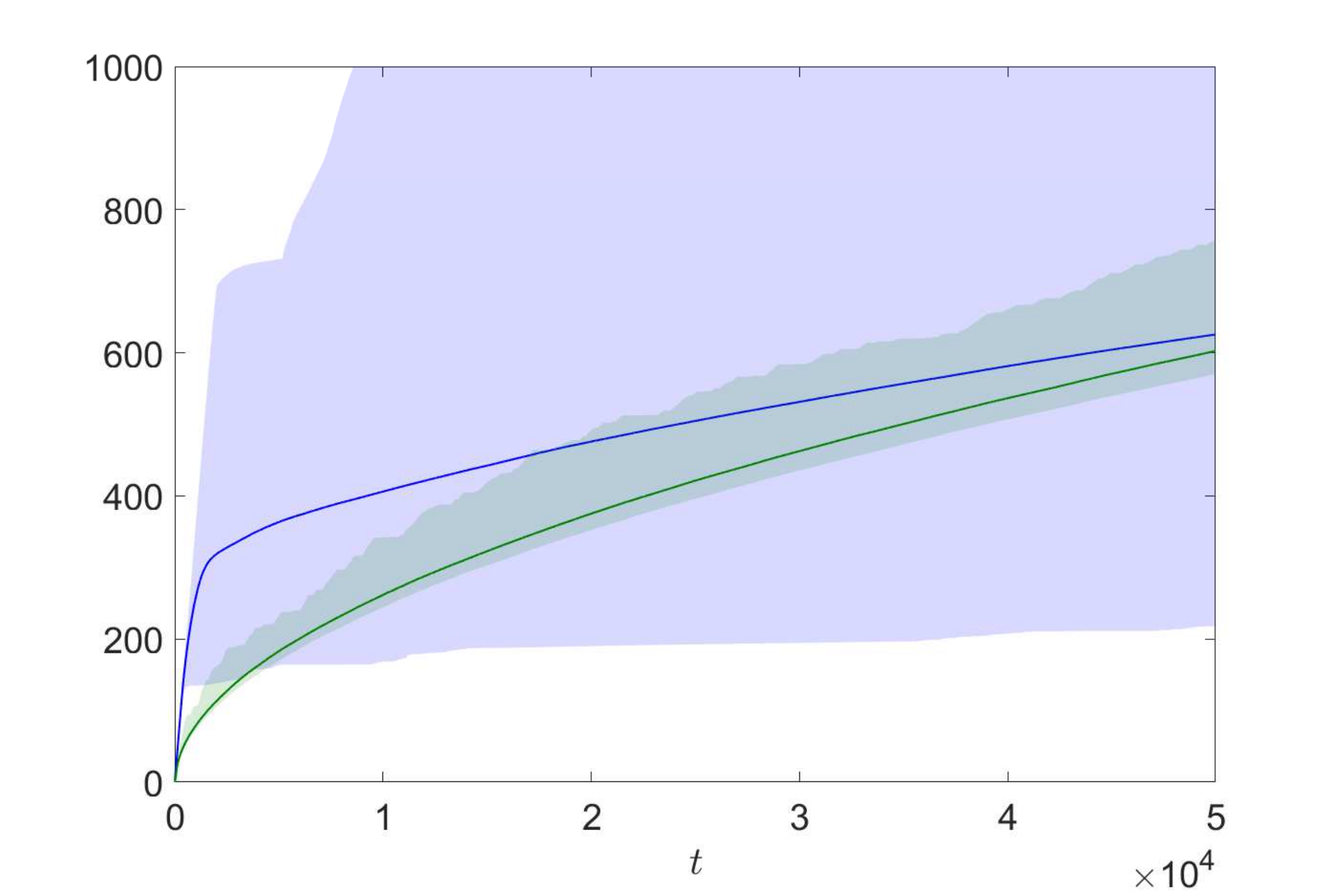}
        \caption{Cumulative regret of the SEGE algorithm (blue) and the CLUCB algorithm (green).}
        \label{fig:regret}
    \end{subfigure}
		    
      \caption{These figures illustrate the empirical performance of the SEGE and CLUCB algorithms.  The solid lines depict empirical means and the shaded regions depict empirical ranges computed from  $250$ independent simulations.}\label{fig:all}
\end{figure*}

\section{Simulation Results}   \label{sec:numerical}
In this section, we conduct a simple numerical study to illustrate the qualitative features of the SEGE algorithm and compare it with the CLUCB algorithm introduced by \cite{kazerouni2017conservative}.

\subsection{Simulation Setup}

\subsubsection{Model Parameters.} We  consider a linear bandit with a two-dimensional input space ($d=2$), and restrict the set of allowable arms $\mathcal{X}$ to be closed disk of radius $r =1$ centered at $\bar{x} = (1,1)$. The true reward parameter is taken to be $\theta^* = (0.6,0.8)$, and the upper bound on its norm is set to $S=1$. We select a baseline arm at random from the set of allowable arms  as $\xs = (1.2,1.9)$, and set the  baseline expected reward  to $b_0= \inprod{\xs,\theta^*}=2.24$. We set the safety threshold to $b=0.8 \times b_0$. The observation noise process $\{\eta_t\}_{t=1}^\infty$ is assumed to be an IID sequence of  zero-mean Normal random variables with standard deviation $\sigma_\eta=1$.

 \subsubsection{SEGE Algorithm.}  We set the parameters of the SEGE algorithm  to $c=0.5$, $\lambda=0.1$, and  $\rho=\bar{\rho}=0.224$. We generate the random exploration process according to $U_t = \bar{x} + \zeta_t$, where $\{\zeta_t\}_{t=1}^\infty$ is a sequence of IID random variables that are uniformly distributed on the unit circle. To enable a direct comparison between the SEGE and CLUCB algorithms, we restrict our attention to a summable sequence of  risk levels that satisfy the conditions of Corollary \ref{cor:kazerouni}. Specifically, we set the sequence of risk levels to $\sprob_t = 6\overline{\sprob}/(\pi^2 t^2)$ for all stages $t\geq 1$, where $\overline{\sprob} = 0.1$.

 \subsubsection{CLUCB Algorithm.}  We note that the implementation of the  CLUCB algorithm requires the repeated solution of a non-convex optimization problem in order to compute UCB arms. To circumvent this intractable calculation, we approximate the continuous set of arms $\mathcal{X}$ by a finite set of arms $\widehat{\mathcal{X}}$ that correspond to a uniform discretization of the boundary of $\mathcal{X}$. The error induced by this approximation is negligible, as  $\max_{x\in\mathcal{X}} \inprod{x,\theta^*}-\max_{x\in\widehat{\mathcal{X}}} \inprod{x,\theta^*} \leq 3 \times 10^{-3}$.

\begin{figure}[htb!]
    \centering
\includegraphics[width=0.7\columnwidth]{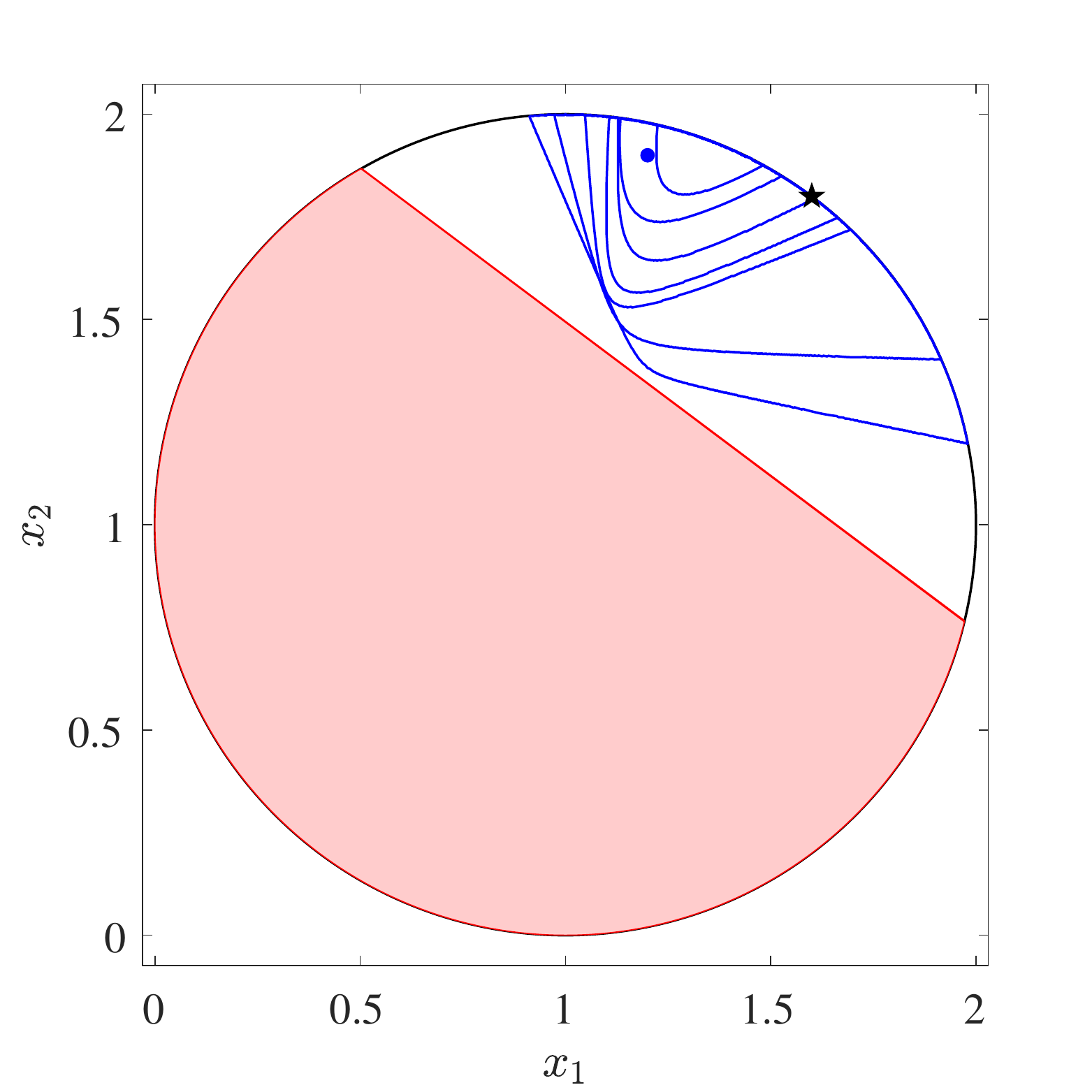}
    \caption{The blue curves depict the gradual expansion of the set of safe arms  $\{x \in \mathcal{X} \ | \ \lcb_t(x)\geq b\}$ over time under the SEGE algorithm for  $t=250$, $500$, $1000$, $2000$, $5000$, $10000$, and $50000$. The blue dot depicts the baseline arm $\xs$,  the black star depicts the optimal arm $X^*$, and the red shaded area depicts the set of unsafe arms.}\label{fig:safe_sets}
\end{figure}

\subsection{Performance of the SEGE Algorithm}
We first discuss the transient behavior and performance of the SEGE algorithm.  As one might expect, the SEGE algorithm initially relies on the baseline arm for safe exploration as depicted in Figure \ref{fig:all}\subref{fig:sege}. Over time, as the algorithm accumulates information, it is able to gradually  expand the set of safe arms as shown in Figure \ref{fig:safe_sets}. This expansion  enables the algorithm to increase the stagewise efficiency with which it safely explores by selecting arms in the vicinity of the safe arm with the largest lower confidence bounds on their expected rewards. In turn, the SEGE algorithm is able to exploit the  information gained to play the greedy with increasing frequency over time. As a result, the growth rate of regret diminishes over time as depicted in Figure \ref{fig:all}\subref{fig:regret}.     
Critically,  Figure \ref{fig:all}\subref{fig:sege} also shows that the SEGE algorithm maintains stagewise safety throughout each of the $250$ independent experiments.

\subsection{Comparison with the CLUCB Algorithm} \label{sec:comp}
Unlike the SEGE algorithm, the CLUCB algorithm is shown to violate the stagewise safety constraint at an early stage in the learning process as depicted in Figure \ref{fig:all}\subref{fig:clucb}. The violation of the stagewise safety constraint by the CLUCB algorithm is not surprising as it is only guaranteed to respect the conservative performance constraint \eqref{eq:conservative const}. The SEGE algorithm, on the other hand, is guaranteed to satisfy the conservative performance constraint, in addition to being stagewise safe (cf. Corollary \ref{cor:kazerouni}).
However, as one might expect, the more stringent  safety guarantee of the SEGE algorithm comes at a cost. Specifically, the regret under the SEGE algorithm initially grows more rapidly than the regret incurred by the CLUCB algorithm, as shown in Figure \ref{fig:all}\subref{fig:regret}.
 However, over time the growth rate of regret of the SEGE algorithm slows down as information accumulates and the need for safe exploration diminishes enabling the algorithm to play the greedy arm more frequently.

\section*{Acknowledgments}
This material is based upon work supported by  the Holland Sustainability Project Trust, and the National Science Foundation under grant no. ECCS-135162 and  IIP-1632124.

\fontsize{9.0pt}{10.0pt} \selectfont
\bibliography{references}
\bibliographystyle{aaai}

\onecolumn
\normalsize

\makeatletter
\def\@seccntformat#1{%
  \expandafter\ifx\csname c@#1\endcsname\c@section\else
  \csname the#1\endcsname\quad
  \fi}
\makeatother
\begin{appendices}
\section{Appendices}
In this Section, we provide a detailed proof of the theoretical results including Lemma \ref{lem:rho}, and Theorems \ref{thm:safety} and \ref{thm:regret}.

\subsection{Proof of Lemma \ref{lem:rho}}
Recall from the definition of $\xS_t$ that $\prob{\inprod{\xS_t,\theta^*}\geq b_0}\geq 1-\sprob_t$. Thus, with probability $1-\sprob_t$, it holds that,
\begin{align}
\inprod{\xse_t,\theta^*}&=\inprod{(1-\rho)\xS_t +\rho U_t,\theta^*}\nonumber\\
&=\inprod{(1-\rho)\xS_t +\rho \bar{x}+\rho H^{1/2} \zeta_t,\theta^*}\nonumber\\
&=\inprod{\xS_t,\theta^*} -\rho\inprod{\xS_t -\bar{x},\theta^*}+\rho\inprod{H^{1/2}\zeta_t,\theta^*}\nonumber\\
&\geq b_0 - \rho\norm{\xS_t -\bar{x}}  \norm{\theta^*}  - \rho \sqrt{\lambda_{\max}(H)}\norm{\theta^*} \label{eq:step 1},
\end{align}
where the inequality follows from the Cauchy-Schwarz inequality and the fact that $\norm{\zeta_t}=1$. For any $x\in \mathcal{X}$, it holds that
\begin{align}
\norm{x-\bar{x}} &\leq \norm{x-\bar{x}}_{H^{-1}} \sqrt{\lambda_{\max}(H)} \nonumber\\
&\leq \sqrt{\lambda_{\max}(H)}, \label{eq:step 2}
\end{align}
where the inequality follows from the definition of $\mathcal{X}$ in Equation \eqref{eq:ellipsoid}. By applying Inequality \eqref{eq:step 2} to Inequality \eqref{eq:step 1}, with probability $1-\sprob_t$, it holds that
\begin{align*}
\inprod{\xse_t,\theta^*}&\geq b_0 - 2\rho \sqrt{\lambda_{\max}(H)}\norm{\theta^*}.
\end{align*}
Recall Assumption \ref{ass:bound_theta} that $\norm{\theta^*}\leq S$. Thus, in order to guarantee that $\prob{\inprod{\xse_t,\theta^*}\geq b}\geq 1-\sprob_t$ it suffices to choose $\rho$ such that
\begin{align}
\rho \leq  \frac{b_0-b }{2S\sqrt{\lambda_{\max}(H)}}.
\end{align}

\subsection{Proof of Theorem \ref{thm:safety}}\label{app:safety}
From Lemma \ref{lem:rho}, it follows that the safe exploration arm $\xse_t$ is $(\sprob_t,b)$-safe by construction. Moreover, under the SEGE algorithm the greedy arm $\xg_t$ is only played if $\lcb_t(\xg_t)\geq b$, which, in turn, implies
\begin{align*}
\prob{\inprod{\xg_t,\theta^*}\geq b}\geq 1-\sprob_t.
\end{align*}
Thus, the greedy arm $\xg_t$ if played is $(\sprob_t,b)$-safe.

\subsection{Proof of Theorem \ref{thm:regret} }\label{app:regret}
As mentioned in the Theoretical Results section, in order to establish an upper bound on the expected regret, we rely on intermediary results. More precisely, to upper bound the expected regret during the greedy exploitation stages, we establish a bound on the stagewise regret under the greedy arm in terms of the mean squared estimation error in Lemma \ref{lem:ce lower}. 
\begin{lemm}[Stagewise Regret] \label{lem:ce lower} The stagewise expected reward under the greedy (certainty equivalent) arm $\xg_t$ is almost surely lower bounded as
\begin{align}
\inprod{\xg_t,\theta^*}\geq \inprod{X^*,\theta^*} - k_1 \lnorm{\theta^* - \widehat{\theta}_{t-1}}^2, \label{eq:ce lower}
\end{align}
for all $t\geq 1$, where the constant $k_1$ is given by
\begin{align*}
k_1= \frac{2\norm{\xs}\lambda_{\max}(H)}{b_0 \sqrt{\lambda_{\min}(H)}}.
\end{align*}
\end{lemm}
The proof of Lemma \ref{lem:ce lower} is postponed to Appendix \ref{app:ce lower}.

Moreover, to upper bound the expected regret during the safe exploitation stages, we establish an upper bound on the expected number of safe exploration stages in Theorem \ref{thm:N_t}. More precisely, let $N_t$ be the number of stages in which a safe exploration arm is played among the first $t$ stages. Then, Theorem \ref{thm:N_t} establishes an upper bound on $\expe{N_t}$ under the SEGE algorithm.

\begin{theore}[Safe Exploration Stages] \label{thm:N_t}
Let  $\{\sprob_t\}_{t=1}^\infty$ be any sequence of risk levels satisfying Inequality \eqref{eq:risk assumption} for all $t\geq 1$. There exists a finite positive constant $C_0$ such that under the SEGE Algorithm \ref{alg:safe alg}, it holds that
\begin{align*}
\expe{N_t}\leq C_0\sqrt{t},
\end{align*} 
for all $t\geq 1$.
\end{theore}
The proof of Lemma \ref{thm:N_t} is postponed to Appendix \ref{app:N_t}.

Recall the definition of expected regret $R_T$
\begin{align*}
R_T = \expe{\sum_{t=1}^T \inprod{X^*-X_t,\theta^*}}. 
\end{align*}
We bound the expected regret in the exploration and exploitation stages separately. Let $\NT^{\mathsf{SE}}$ ($\NT^{\mathsf{CE}}$) be the stages in which a safe exploration arm (greedy arm) is played in the first $T$ stages. The expected regret can be decomposed into two parts,
\begin{align}
R_T &= \expe{\sum_{t\in \NT^{\mathsf{SE}}} \inprod{X^*-\xse_t,\theta^*}}  + \expe{\sum_{t\in \NT^{\mathsf{CE}}} \inprod{X^*-\xg_t,\theta^*}}. \label{eq:regret decomp}
\end{align}

\

\quad \textit{The regret in safe exploration stages}:\  From the fact that $\norm{x}\leq L$ for all $x\in \mathcal{X}$, $\norm{\theta^*}\leq S$, and the Cauchy-Schwarz inequality it almost surely holds that
\begin{align*}
\sum_{t\in \NT^{\mathsf{SE}}} \inprod{X^*-\xse_t,\theta^*}  \leq 2LS N_T.
\end{align*} 
From Theorem \ref{thm:N_t}, it follows that
\begin{align*}
\expe{\sum_{t\in \NT^{\mathsf{SE}}} \inprod{X^*-\xse_t,\theta^*}} \leq 2LS \expe{N_T}\leq 2LSC_0 \sqrt{T}.
\end{align*}

\

\quad \textit{The regret in greedy exploitation stages}:\ Recall that the greedy arm is only played if $\lcb_t(\xg_t)\geq b$ and $\lambda_{\min}(V_t)\geq c\sqrt{t}$. Notice that it almost surely holds that $\inprod{x,\theta^*} \leq LS$ for all $x\in\mathcal{X}$.   Then, for all $t\in \NT^{\mathsf{CE}}$, it holds that
\begin{align*}
\expe{\inprod{X^*-\xg_t,\theta^*} } &= \int_{0}^{2LS}\prob{\left\{ \inprod{X^*-\xg_t,\theta^*}\geq \gamma\right\} \bigcap\left\{ \lambda_{\min}(V_{t-1})\geq c\sqrt{t} \right\} \bigcap\left\{ \lcb_t(\xg_t)\geq b \right\}   }\ d\gamma\\
&\leq \int_{0}^{2LS}\prob{\left\{ \inprod{X^*-\xg_t,\theta^*}\geq \gamma\right\} \bigcap\left\{ \lambda_{\min}(V_{t-1})\geq c\sqrt{t} \right\}   }\ d\gamma\\
&\leq  \int_{0}^{2LS}\prob{ \left\{ k_1\norm{\theta^* - \widehat{\theta}_{t-1}}^2\geq \gamma\right\} \bigcap\left\{ \lambda_{\min}(V_{t-1})\geq c\sqrt{t} \right\} }\ d\gamma,
\end{align*}
where the last inequality follows from the bound on the stagewise regret in Inequality \eqref{eq:ce lower}. From the Cauchy-Schwarz inequality it follows that $\norm{\theta^* - \widehat{\theta}_{t-1}}_{V_{t-1}}\geq \norm{\theta^* - \widehat{\theta}_{t-1}} \sqrt{\lambda_{\min}(V_{t-1})}$. Then,
\begin{align}
\expe{\inprod{X^*-\xg_t,\theta^*} }&\leq \int_{0}^{2LS}\prob{ \left\{ \norm{\theta^* - \widehat{\theta}_{t-1}}^2_{V_{t-1}}\geq \frac{\gamma c\sqrt{t}}{k_1}\right\} \bigcap \left\{\lambda_{\min}(V_{t-1})\geq c\sqrt{t}\right\} }\ d\gamma \nonumber\\
&\leq \int_{0}^{2LS}\prob{ \norm{\theta^* - \widehat{\theta}_{t-1}}^2_{V_{t-1}}\geq \frac{\gamma c\sqrt{t}}{k_1}  }\ d\gamma \nonumber\\
&\leq  \gamma_{t-1}  +\int_{ \gamma_{t-1}}^{2LS} \prob{ \norm{\theta^* - \widehat{\theta}_{t-1}}^2_{V_{t-1}}\geq \frac{\gamma c\sqrt{t}}{k_1} }\ d\gamma, \label{eq:integral prob}
\end{align}
where the last inequality follows from upper bounding the integrand by $1$ for $\gamma\leq \gamma_{t-1}$. Here, $ \gamma_{t-1}$ is defined as
\begin{equation}
\gamma_{t-1} =  k_7 \frac{\log\left(t\right)}{\sqrt{t}}, \label{eq:gamma s}
\end{equation}
where $k_4$ is defined as
\begin{align}
k_7 = \frac{2k_1 d\sigma_{\eta}^2}{c} \left(\log(1+L^2/\lambda) + \frac{2\lambda S^2}{d\sigma_{\eta}^2}\right). \label{eq:k_7}
\end{align}
Recall from the definition of $\mathcal{C}_{t-1}(\delta)$ that for any $\delta\in(0,1)$, it holds that
\begin{align}
&\prob{\norm{\theta^* - \widehat{\theta}_{t-1}}_{V_{t-1}}\geq r_t(\delta) } \leq  \delta  . \label{eq:error prob int 1}
\end{align}
In order to apply Inequality \eqref{eq:error prob int 1} to \eqref{eq:integral prob}, we set $\delta$ such that $r_t^2(\delta) = c\sqrt{t} \gamma/k_1 $, i.e.,
\begin{align*}
\delta = (1+tL^2/\lambda)\exp\left(-\frac{1}{d\sigma_{\eta}^2}\left(\sqrt{\frac{c\sqrt{t}\gamma}{k_1}}-\sqrt{\lambda}S\right)^2\right). 
\end{align*}
Using the fact that for any two real numbers $x,y>0$, we have that $(\sqrt{x}-\sqrt{y})^2\geq x/2 -  y$, we get
\begin{align}
&\delta \leq  (1+tL^2/\lambda)\exp\left( \frac{2\lambda S^2}{d\sigma_{\eta}^2}\right)\exp\left(-\frac{c\gamma}{2k_1 d\sigma_{\eta}^2}\sqrt{t}\right).\label{eq:delta part 1}
\end{align}
Then, by applying Inequality \eqref{eq:error prob int 1} and \eqref{eq:delta part 1} to Inequality \eqref{eq:integral prob}, we get
\begin{align}
\expe{\inprod{X^*-\xg_t,\theta^*} }&\leq \gamma_{t-1}  + (1+tL^2/\lambda)\exp\left( \frac{2\lambda S^2}{d\sigma_{\eta}^2}\right) \int_{ \gamma_{t-1}}^{2LS} \exp\left(-\frac{c\gamma}{2k_1 d\sigma_{\eta}^2}\sqrt{t}\right)\ d\gamma \nonumber\\
&\leq  \gamma_{t-1}  + (1+tL^2/\lambda)\exp\left( \frac{2\lambda S^2}{d\sigma_{\eta}^2}\right)\exp\left(-\frac{c\gamma_{t-1}}{2k_1 d\sigma_{\eta}^2}\sqrt{t}\right)  \frac{2k_1 d\sigma_{\eta}^2}{c\gamma \sqrt{t}}\nonumber\\
&\leq \gamma_{t-1}  +   \frac{2k_1 d\sigma_{\eta}^2}{c \sqrt{t}},\label{eq:delta part 2}
\end{align}
where the last inequality follows from the fact that $(1+tL^2/\lambda)\exp\left( \frac{2\lambda S^2}{d\sigma_{\eta}^2}\right)\exp\left(-\frac{c\gamma_{t-1}}{2k_1 d\sigma_{\eta}^2}\sqrt{t}\right)\leq 1$ by definition, i.e.,
\begin{align*}
(1+tL^2/\lambda)\exp\left( \frac{2\lambda S^2}{d\sigma_{\eta}^2}\right)\exp\left(-\frac{c\gamma_{t-1}}{2k_1 d\sigma_{\eta}^2}\sqrt{t}\right)&\leq t(1+L^2/\lambda)\exp\left( \frac{2\lambda S^2}{d\sigma_{\eta}^2}\right)\exp\left(-\frac{c k_7 \log(t)}{2k_1 d\sigma_{\eta}^2}\right)\\
&= (1+L^2/\lambda)\exp\left( \frac{2\lambda S^2}{d\sigma_{\eta}^2}\right)\exp\left(-\frac{c k_7  }{2k_1 d\sigma_{\eta}^2}\right)\\
&=1,
\end{align*}
where the last inequality follows from the definition of $k_7$ in inequality \eqref{eq:k_7}. Thus, using Inequality \eqref{eq:delta part 2} and the definition of $\gamma_{t-1}$ in \eqref{eq:gamma s}, we get
\begin{align*}
\expe{\sum_{t\in \NT^{\mathsf{CE}}} \inprod{X^*-\xg_t,\theta^*}} &\leq \sum_{t=2}^T \left(k_7 \frac{\log\left(t\right)}{\sqrt{t}} + \frac{2k_1 d\sigma_{\eta}^2}{c \sqrt{t}} \right)\\
&\leq \sum_{t=2}^T \left(k_7 \frac{\log(T)}{\sqrt{t}} + \frac{2k_1 d\sigma_{\eta}^2}{c \sqrt{t}} \right)\\
&\leq \left(k_7 \log(T) + \frac{2k_1 d\sigma_{\eta}^2}{c  } \right) \int_{t=1}^T \frac{1}{\sqrt{t}}\ dt\\
&\leq C_0  \sqrt{T}+ C_1 \sqrt{T}\log(T),
\end{align*}
where $C_2$ and $C_3$ are defined as
\begin{align*}
C_2 &=  \frac{4k_1 d\sigma_{\eta}^2}{c  }, \\
C_3 &= 2k_7.
\end{align*}
Thus, by defining $C = 2LS C_0 + C_2 + C_3$, we get the desired upper bound on regret.

\subsection{Proof of Theorem \ref{thm:N_t}}\label{app:N_t}

From the definition of $N_t$, it follows that $N_0=0$ and for all $t\geq 0$, 
\begin{align}
N_{t+1} = \begin{cases}
N_t,& \lambda_{\min}(V_t)\geq c\sqrt{t} \ \text{and}\ \lcb_t(\xg_t)\geq b,\\
N_t+1,& \text{otherwise}. \label{eq:N_t}
\end{cases}
\end{align}

Fix $\mu\in (0,1)$. Define the random process $\{Z\}_{t=1}^\infty$ as follows. For any $t\geq 0$, $Z_t$ is defined as
\begin{align}
Z_t = 0 \vee \left(N_t - \left\lceil \frac{c\sqrt{t}}{\mu \rho^2\sigma^2}\right\rceil\right). \label{eq:Z_t}
\end{align}
Notice that if $N_t < \left\lceil \frac{c\sqrt{t}}{\mu \rho^2\sigma^2}\right\rceil$ then $Z_{t+1} = 0$. Thus,
\begin{align}
\expe{Z_{t+1}} &= \expe{Z_{t}} + \expe{ \mathds{1}\left\{\left\{N_t \geq  \left\lceil \frac{c\sqrt{t}}{\mu \rho^2\sigma^2}\right\rceil\right\} \cap \left(\left\{\lambda_{\min}(V_t)\leq c\sqrt{t}\right\}\cup \left\{{\normalfont{\lcb}_t}(\xg_t)\leq b\right\}\right) \right\}} \nonumber\\
&= \expe{Z_{t}} + \prob{ \left\{N_t \geq  \left\lceil \frac{c\sqrt{t}}{\mu \rho^2\sigma^2}\right\rceil\right\} \cap \left(\left\{\lambda_{\min}(V_t)\leq c\sqrt{t}\right\}\cup \left\{{\normalfont{\lcb}_t}(\xg_t)\leq b\right\}\right) }\nonumber\\
&=\sum_{k=1}^t \prob{ \left\{N_k \geq  \left\lceil \frac{c\sqrt{k}}{\mu \rho^2\sigma^2}\right\rceil\right\} \cap \left(\left\{\lambda_{\min}(V_k)\leq c\sqrt{k}\right\}\cup \left\{{\normalfont{\lcb}_k}(\xg_k)\leq b\right\}\right) } \label{eq:event prob}
\end{align}
To establish an upper bound on the probability of the events in Equation \eqref{eq:event prob}, we establish the following relationship.
\begin{lemm} \label{lem:relaxation}
Conditioned on the event $\lambda_{\min}(V_t)\geq c\sqrt{t}$, we have that $ \theta^*\in \mathcal{C}_{t-1}\left(\sprob_t\wedge\widehat{\sprob}_t\right)$ implies ${\normalfont{\lcb}_t}(\xg_t)\geq b$, i.e., 
\begin{align*}
\left\{\lambda_{\min}(V_t)\geq c\sqrt{t}\right\}\cap \left\{\theta^*\in \mathcal{C}_{t-1}\left(\sprob_t\wedge\widehat{\sprob}_t\right)\right\}  \subseteq  \left\{\lambda_{\min}(V_t)\geq c\sqrt{t}\right\}\cap \left\{{\normalfont{\lcb}_t}(\xg_t)\geq b\right\} .
\end{align*}
 Here, $\widehat{\sprob}_t$ is defined as
\begin{align*}
\widehat{\sprob}_t= \left(1+\frac{tL^2}{\lambda}\right)\exp\left(-\frac{k_2c\sqrt{t}-2\lambda S^2}{2\sigma_\eta^2 d}\right),
\end{align*}
where the positive constant $k_2$ is defined as
\begin{align*}
k_2=\left(\sqrt{\frac{b_0-b}{k_1}+\frac{L^2}{k_1^2}}-\frac{L}{k_1}\right)^2
\end{align*}
\end{lemm}
By applying De Morgan's law and Lemma \ref{lem:relaxation} to Inequality \eqref{eq:event prob}, we get
\begin{align*}
\expe{Z_{t+1}}&\leq \sum_{k=1}^t \prob{ \left\{N_k \geq  \left\lceil \frac{c\sqrt{k}}{\mu \rho^2\sigma^2}\right\rceil\right\} \cap \left(\left\{\lambda_{\min}(V_k)\leq c\sqrt{k}\right\}\cup \left\{\theta^*\notin\mathcal{C}_{k-1}\left(\sprob_k\wedge\widehat{\sprob}_k\right)\right\}\right) }\\
&\leq \sum_{k=1}^t \prob{ \left\{N_k \geq  \left\lceil \frac{c\sqrt{k}}{\mu \rho^2\sigma^2}\right\rceil\right\} \cap  \left\{\lambda_{\min}(V_k)\leq c\sqrt{k}\right\}} + \prob{\theta^*\notin\mathcal{C}_{k-1}\left(\sprob_k\wedge\widehat{\sprob}_k\right)}\\
&\leq \sum_{k=1}^t \prob{ \left\{N_k \geq  \left\lceil \frac{c\sqrt{k}}{\mu \rho^2\sigma^2}\right\rceil\right\} \cap  \left\{\lambda_{\min}(V_k)\leq c\sqrt{k}\right\}} +  \sprob_k\wedge\widehat{\sprob}_k \\
&\leq \sum_{k=1}^t \prob{ \left\{N_k \geq  \left\lceil \frac{c\sqrt{k}}{\mu \rho^2\sigma^2}\right\rceil\right\} \cap  \left\{\lambda_{\min}(V_k)\leq c\sqrt{k}\right\}} +  \widehat{\sprob}_k.
\end{align*} 
Using the total probability theorem, we have that
\begin{align}
\prob{ \left\{N_k \geq  \left\lceil \frac{c\sqrt{k}}{\mu \rho^2\sigma^2}\right\rceil\right\}\cap  \left\{\lambda_{\min}(V_k)\leq c\sqrt{k}\right\}}&= \sum_{n=\left\lceil \frac{c\sqrt{k}}{\mu \rho^2\sigma^2}\right\rceil}^\infty \prob{\lambda_{\min}(V_k)\leq c\sqrt{k}\mid N_k = n} \prob{N_k=n}\nonumber\\
&\leq  \sum_{n=\left\lceil \frac{c\sqrt{k}}{\mu \rho^2\sigma^2}\right\rceil}^\infty \prob{\lambda_{\min}(V_k)\leq  \mu \rho^2 \sigma^2 N_k \mid N_k = n} \prob{N_k=n}. \label{eq:sum lmin}
\end{align}

We now establish an upper bound on the minimum eigenvalue of $V_t$ in terms of the number of safe exploration stages.

\begin{lemm}[Random Exploration] \label{lem:lmin}
Under the SEGE  Algorithm, for any $\mu\in(0,1)$ it holds that
\begin{align}
&\prob{\lambda_{\min}(V_t) \leq   \mu \rho^2 \sigma^2 N_t \mid N_t = n } \leq de^{-k_4 (1-\mu)^2 n}, \label{eq:lmin}
\end{align}
where $c_1$ is defined as 
\begin{align*}
k_4 &=  \frac{\rho^4\sigma^4}{2k_3^2},\\
k_3 &= 2\rho((1-\rho)L + \rho \norm{\bar{x}}) \sqrt{\lambda_{\max}(H)}   +  \rho^2 \lambda_{\max}(H)- \rho^2\sigma^2 d.
\end{align*}
\end{lemm}
By applying Inequality \eqref{eq:lmin} to Inequality \eqref{eq:sum lmin}, we 
\begin{align*}
\prob{ \left\{N_k \geq  \left\lceil \frac{c\sqrt{k}}{\mu \rho^2\sigma^2}\right\rceil\right\}\cap  \left\{\lambda_{\min}(V_k)\leq c\sqrt{k}\right\}}&\leq  \sum_{n=\left\lceil \frac{c\sqrt{k}}{\mu \rho^2\sigma^2}\right\rceil}^\infty d \exp\left(-k_4 (1-\mu)^2 n\right)\prob{N_k=n}\\
&\leq d\exp\left(-\frac{k_4 (1-\mu)^2 c}{\mu \rho^2\sigma^2}\sqrt{k}\right) \sum_{n=\left\lceil \frac{c\sqrt{k}}{\mu \rho^2\sigma^2}\right\rceil}^\infty  \prob{N_k=n}\\
&\leq d\exp\left(-\frac{k_4 (1-\mu)^2 c}{\mu \rho^2\sigma^2}\sqrt{k}\right).
\end{align*}
Thus,
\begin{align*}
\expe{Z_{t}}&\leq \sum_{k=1}^{t-1} \left( d\exp\left(-\frac{k_4 (1-\mu)^2 c}{\mu \rho^2\sigma^2}\sqrt{k}\right) + \left(1+\frac{kL^2}{\lambda}\right)\exp\left(-\frac{k_2c\sqrt{k}-2\lambda S^2}{2\sigma_\eta^2 d}\right) \right)\\
&\leq \sum_{k=1}^{t-1} \left( d\exp\left(-\frac{k_4 (1-\mu)^2 c}{\mu \rho^2\sigma^2}\sqrt{k}\right) +  k_5 k \exp\left(-\frac{k_2c}{2\sigma_\eta^2 d}\sqrt{k}\right) \right),
\end{align*}
where $k_5$ is defined as
\begin{align*}
k_5 = \left(1+\frac{L^2}{\lambda}\right)\exp\left(\frac{ \lambda S^2}{ \sigma_\eta^2 d}\right).
\end{align*}
Then,
\begin{align*}
\expe{Z_{t}}&\leq d \int_0^t   \exp\left(-\frac{k_4 (1-\mu)^2 c}{\mu \rho^2\sigma^2}\sqrt{x}\right)\ d x +  k_5 \int_0^t x \exp\left(-\frac{k_2c}{2\sigma_\eta^2 d}\sqrt{x}\right)  \ d x\\
&\leq d \int_0^\infty   \exp\left(-\frac{k_4 (1-\mu)^2 c}{\mu \rho^2\sigma^2}\sqrt{x}\right)\ d x +  k_5 \int_0^\infty x \exp\left(-\frac{k_2c}{2\sigma_\eta^2 d}\sqrt{x}\right)  \ d x\\
&= -d \frac{2(\mu \rho^2\sigma^2)^2}{(k_4 (1-\mu)^2 c)^2} \exp\left(-\frac{k_4 (1-\mu)^2 c}{\mu \rho^2\sigma^2}\sqrt{x}\right) \left(\frac{k_4 (1-\mu)^2 c}{\mu \rho^2\sigma^2}\sqrt{x}+1\right)\bigg|_{0}^\infty \\ 
&\qquad - \frac{2 (2\sigma_\eta^2 d)^4}{ (k_2c)^4} \exp\left(-\frac{k_2c}{2\sigma_\eta^2 d}\sqrt{x}\right) \left( \left(\frac{k_2c}{2\sigma_\eta^2 d}\right)^3 x^{3/2} + 3 \left(\frac{k_2c}{2\sigma_\eta^2 d}\right)^2x + 6\frac{k_2c}{2\sigma_\eta^2 d} \sqrt{x} +6 \right) \bigg|_{0}^\infty\\
&= k_6,
\end{align*}
where $k_6$ is defined as
\begin{align*}
k_6 &= d \frac{2(\mu \rho^2\sigma^2)^2}{(k_4 (1-\mu)^2 c)^2} + k_5\frac{12 (2\sigma_\eta^2 d)^4}{ (k_2c)^4} .
\end{align*}
Thus,
\begin{align*}
\expe{N_t} &\leq \expe{Z_t} + \left\lceil \frac{c\sqrt{t}}{\mu \rho^2\sigma^2}\right\rceil\\
&\leq k_6 + 1 + \frac{c\sqrt{t}}{\mu \rho^2\sigma^2}.
\end{align*}
Setting $C_0 = k_6+1+\frac{c}{\mu \rho^2\sigma^2}$ concludes the proof.

\subsection{Proof of Lemma \ref{lem:ce lower}}\label{app:ce lower}
For each $t\geq 1$, the expected reward under the greedy arm is lower bounded as 
\begin{align*}
 \inprod{\xg_t,\theta^*}  &= \inprod{X^*,\theta^*} - \inprod{X^*-\xg_t,\theta^*}\\
&= \inprod{X^*,\theta^*} - \inprod{X^*-\xg_t,\theta^*-\widehat{\theta}_{t-1}} - \inprod{X^*-\xg_t,\widehat{\theta}_{t-1}}\\
&\geq \inprod{X^*,\theta^*} - \inprod{X^*-\xg_{t},\theta^*-\widehat{\theta}_{t-1}},
\end{align*}
where the inequality follows from the fact that $ \inprod{\xg_t,\widehat{\theta}_{t-1}}\geq \inprod{x,\widehat{\theta}_{t-1}}$ for all $x\in\mathcal{X}$ as the greedy arm is the optimal arm for the reward parameter $\widehat{\theta}_{t-1}$. Using the Cauchy-Schwarz inequality, we get
\begin{align}
\inprod{\xg_t,\theta^*}&\geq \inprod{X^*,\theta^*} - \norm{X^*-\xg_t}  \norm{\theta^*-\widehat{\theta}_{t-1}} \nonumber\\
&= \inprod{X^*,\theta^*} - \lnorm{\frac{H\theta^*}{\norm{\theta^*}_H}-\dfrac{H\widehat{\theta}_{t-1}}{\norm{\widehat{\theta}_{t-1}}_H}} \norm{\theta^*-\widehat{\theta}_{t-1}}. \label{eq:diff reward}
\end{align}
We now show that 
\begin{align}
\lnorm{\frac{H\theta^*}{\norm{\theta^*}_H}-\dfrac{H\widehat{\theta}_{t-1}}{\norm{\widehat{\theta}_{t-1}}_H}}  \leq  \frac{2\norm{\xs} {\lambda_{\max}(H)}}{b_0 \sqrt{\lambda_{\min}(H)} }\norm{\theta^*-\widehat{\theta}_{t-1}} . \label{eq:diff par}
\end{align}
Using the triangle inequality, we get
\begin{align*}
 \lnorm{\frac{H\theta^*}{\norm{\theta^*}_H}-\dfrac{H\widehat{\theta}_{t-1}}{\norm{\widehat{\theta}_{t-1}}_H}}  &= \lnorm{\frac{H\theta^*}{\norm{\theta^*}_H}- \frac{H\widehat{\theta}_{t-1}}{\norm{\theta^*}_H} +\frac{H\widehat{\theta}_{t-1}}{\norm{\theta^*}_H}- \dfrac{H\widehat{\theta}_{t-1}}{\norm{\widehat{\theta}_{t-1}}_H}}\\
&\leq  \lnorm{\frac{H\theta^*}{\norm{\theta^*}_H}- \frac{H\widehat{\theta}_{t-1}}{\norm{\theta^*}_H}} + \lnorm{\frac{H\widehat{\theta}_{t-1}}{\norm{\theta^*}_H}- \dfrac{H\widehat{\theta}_{t-1}}{\norm{\widehat{\theta}_{t-1}}_H}}\\
&= \frac{1}{\norm{\theta^*}_H} \norm{H(\theta^*-\widehat{\theta}_{t-1})} + \frac{\norm{H\widehat{\theta}_{t-1}}}{\norm{\theta^*}_H \norm{\widehat{\theta}_{t-1}}_H} \left|  \norm{\theta^*}_H - \norm{\widehat{\theta}_{t-1}}_H \right|\\
&\leq \frac{1}{\norm{\theta^*}_H} \norm{H(\theta^*-\widehat{\theta}_{t-1})}   + \frac{\norm{H\widehat{\theta}_{t-1}}}{\norm{\theta^*}_H \norm{\widehat{\theta}_{t-1}}_H}   \norm{\theta^* - \widehat{\theta}_{t-1}}_H, 
\end{align*}
where the last inequality follows from the reverse triangle inequality. Using the Cauchy-Schwarz inequality and the fact that $\norm{H} = \lambda_{\max}(H)$, we get
\begin{align}
&\frac{1}{\norm{\theta^*}_H} \norm{H(\theta^*-\widehat{\theta}_{t-1})} + \frac{\norm{H\widehat{\theta}_{t-1}}}{\norm{\theta^*}_H \norm{\widehat{\theta}_{t-1}}_H}   \norm{\theta^* - \widehat{\theta}_{t-1}}_H  \nonumber\\
&\leq  \frac{\lambda_{\max}(H)}{\norm{\theta^*}_H} \norm{\theta^*-\widehat{\theta}_{t-1}}  + \frac{\sqrt{\lambda_{\max}(H)}\norm{\widehat{\theta}_{t-1}}_H}{\norm{\theta^*}_H \norm{\widehat{\theta}_{t-1}}_H}   \sqrt{\lambda_{\max}(H)}\norm{\theta^* - \widehat{\theta}_{t-1}}\nonumber\\
&=\frac{2\lambda_{\max}(H)}{\norm{\theta^*}_H} \norm{\theta^* - \widehat{\theta}_{t-1}}, \label{eq:norm H theta}
\end{align}
where the inequality follows from the fact that $\norm{H\widehat{\theta}_{t-1}}\leq \norm{H^{1/2}}\norm{H^{1/2}\widehat{\theta}_{t-1}} = \sqrt{\lambda_{\max}(H)}\norm{\widehat{\theta}_{t-1}}_H$. Recall from Assumption \ref{ass:baseline} that $\inprod{\xs,\theta^*}\geq b_0$. So, $\norm{\xs}\norm{\theta^*}\geq b_0$
\begin{align}
\norm{\theta^*}_H\geq \frac{\sqrt{\lambda_{\min}(H)} b_0}{\norm{\xs}}. \label{eq:norm H theta 2}
\end{align} 
Thus, by applying Inequality \eqref{eq:norm H theta 2} to \eqref{eq:norm H theta}, we get Inequality \eqref{eq:diff par}. Finally, combining Inequalities \eqref{eq:diff reward} and \eqref{eq:diff par} yields the desired lower bound on the expected reward of the greedy arm.

\subsection{Proof of Lemma \ref{lem:relaxation}}\label{app:relaxation}

For $x\in \mathcal{X}$, let $\widehat{\lcb}_t(x)$ be the lower confidence bound on the reward of arm $x$ computed using the confidence set $\mathcal{C}_{t-1}(\widehat{\sprob}_t)$, i.e.,
\begin{align*}
\widehat{\lcb}_t(x) = \inprod{x,\widehat{\theta}_{t-1}} - r_t(\widehat{\sprob}_t) \norm{x}_{V_{t-1}^{-1}}.
\end{align*}
Notice that for any $\delta,\widehat{\delta}\in(0,1)$ the condition $\widehat{\delta}\leq \delta$ implies that $r_t(\widehat{\delta})\geq r_t(\delta)$. Thus, if $\widehat{\sprob}_t\leq \sprob_t$ then $\widehat{\lcb}_t(x)\leq \lcb_t(x)$.

\

Recall that $\theta^*\in \mathcal{C}_{t-1}(\sprob_t)$ implies that ${\normalfont{\lcb}_t}(\xg_t)\geq b$. Thus, to prove the Lemma it suffices to show that given $ \lambda_{\min}(V_t)\geq c\sqrt{t}$ and $\theta^*\in \mathcal{C}_{t-1}(\widehat{\sprob}_t)$ , we have that $\widehat{\lcb}_t(\xg_t)\geq b$. It holds that
\begin{align*}
\widehat{\lcb}_t(\xg_t) &= \inprod{\xg_t,\widehat{\theta}_{t-1}} - r_t(\widehat{\sprob}_t) \lnorm{\xg_t}_{V_{t-1}^{-1}}\\
&\geq \inprod{\xg_t,\theta^*} - 2r_t(\widehat{\sprob}_t)\lnorm{\xg_t}_{V_{t-1}^{-1}}\\
&\geq \inprod{X^*,\theta^*} - k_1 \lnorm{\theta^*-\widehat{\theta}_{t-1}}^2 - 2r_t(\widehat{\sprob}_t)\lnorm{\xg_t}_{V_{t-1}^{-1}},
\end{align*}
where the first inequality follows from the fact that $\norm{\theta^* -\widehat{\theta}_{t-1} }_{V_{t-1}}\leq r_t(\widehat{\sprob}_t)$ and the second inequality follows from Lemma \ref{lem:ce lower}.
Then, using the fact that $\inprod{X^*,\theta^*}\geq b_0$ we get
\begin{align*}
\widehat{\lcb}_t(\xg_t) &\geq b_0 - k_1 \lnorm{\theta^*-\widehat{\theta}_{t-1}}^2 - 2r_t(\widehat{\sprob}_t)\lnorm{\xg_t}_{V_{t-1}^{-1}}\\
&\geq b_0 - \frac{k_1}{\lambda_{\min}(V_{t-1})} \lnorm{\theta^*-\widehat{\theta}_{t-1}}^2_{V_{t-1}} - 2r_t(\widehat{\sprob}_t)\lnorm{\xg_t}_{V_{t-1}^{-1}}\\
&\geq b_0 - k_1\frac{r_t(\widehat{\sprob}_t)^2}{\lambda_{\min}(V_{t-1})}  - 2L\frac{r_t(\widehat{\sprob}_t)}{\sqrt{\lambda_{\min}(V_{t-1})}} \\
&\geq b_0 - k_1\frac{r_t(\widehat{\sprob}_t)^2}{\lambda_{\min}(V_{t-1})}  - 2L\frac{r_t(\widehat{\sprob}_t)}{\sqrt{\lambda_{\min}(V_{t-1})}} \\
&= b_0+ \frac{L^2}{k_1} - k_1\left(\frac{r_t(\widehat{\sprob}_t)}{\sqrt{\lambda_{\min}(V_{t-1})}} +\frac{L}{k_1} \right)^2\\
&\geq  b_0+ \frac{L^2}{k_1} - k_1\left(\frac{r_t(\widehat{\sprob}_t)}{\sqrt{c\sqrt{t}}} +\frac{L}{k_1} \right)^2.
\end{align*}
In order to show that $\widehat{\lcb}_t(\xg_t)\geq b$, it suffices to show that
\begin{align*}
 b_0+ \frac{L^2}{k_1} - k_1\left(\frac{r_t(\widehat{\sprob}_t)}{\sqrt{c\sqrt{t}}} +\frac{L}{k_1} \right)^2\geq b,
\end{align*}
or equivalently
\begin{align*}
r_t(\widehat{\sprob}_t)^2 \leq  \left(\sqrt{\frac{b_0-b}{k_1}+\frac{L^2}{k_1^2}}-\frac{L}{k_1}\right)^2 c\sqrt{t}. 
\end{align*}
It holds that
\begin{align*}
r_t(\delta)^2\leq 2\sigma_\eta^2 d \log\left(\frac{1+tL^2/\lambda}{\delta}\right)+2\lambda S^2.
\end{align*}
From the definition of $\widehat{\sprob}_t$ it immediately follows that
\begin{align*}
2\sigma_\eta^2 d \log\left(\frac{1+tL^2/\lambda}{\widehat{\sprob}_t}\right)+2\lambda S^2 = \left(\sqrt{\frac{b_0-b}{k_1}+\frac{L^2}{k_1^2}}-\frac{L}{k_1}\right)^2 c\sqrt{t}.
\end{align*}
Thus, if  $\widehat{\sprob}_t\leq \sprob_t$ then  $\lcb_t(\xg_t)\geq b$.

\subsection{Proof of Lemma \ref{lem:lmin}} \label{app:lmin}
Let $\mathcal{N}_t^{\mathsf{SE}}$ be the set of stages in which a safe exploration arm is played up to and including stage $t$. Our objective is to establish a lower bound on the minimum eigenvalue of $V_t$ in terms on $N_t$. As $y y^\top$ is a positive semidefinite matrix for any $y\in \mathbb{R}^d$, it holds that 
\begin{align*}
V_t&=\lambda I + \sum_{k=1}^t X_t X_t^\top\\
&\succeq  \sum_{k\in \mathcal{N}_t^{\mathsf{SE}}} \xse_k{\xse_k}^\top\\
&=  \sum_{k\in \mathcal{N}_t^{\mathsf{SE}}} \bigg(((1-\rho)\xS_k + \rho \bar{x} + \rho H^{1/2}\zeta_k) ((1-\rho)\xS_k + \rho \bar{x} + \rho H^{1/2}\zeta_k)^\top\bigg)\\
&\succeq \sum_{k\in \mathcal{N}_t^{\mathsf{SE}}} \bigg(((1-\rho)\xS_k + \rho \bar{x})( \rho H^{1/2}\zeta_k)^\top  + \rho H^{1/2}\zeta_k ((1-\rho)\xS_k + \rho \bar{x})^\top  +  \rho^2 H^{1/2}\zeta_k \zeta_k^\top H^{1/2}\bigg)\\
& =  \sum_{k\in \mathcal{N}_t^{\mathsf{SE}}} \left( \rho^2  H^{1/2}\expe{ \zeta_k \zeta_k^\top}H^{1/2}  + W_k\right)\\
&\succeq  \sum_{k\in \mathcal{N}_t^{\mathsf{SE}}} \left( \rho^2 \lambda_{\min}\left(H^{1/2}\expe{ \zeta_k \zeta_k^\top}H^{1/2}\right) + W_k\right),
\end{align*}
where $W_k$ is defined as
\begin{align}
W_k = & ((1-\rho)\xS_k + \rho \bar{x})(\rho H^{1/2}\zeta_k)^\top  + \rho H^{1/2}\zeta_k ((1-\rho)\xS_k + \rho \bar{x})^\top   +  \rho^2 H^{1/2}(\zeta_k \zeta_k^\top  -\expe{ \zeta_k \zeta_k^\top}) H^{1/2} . \label{eq:w_k}
\end{align}
Recall that $\sigma^2$ is defined as the minimum eigenvalue of the covariance matrix of $U_k$, i.e.,
\begin{align*}
\sigma^2&= \lambda_{\min}\left(\expe{\left(U_k-\expe{U_k}\right)\left( U_k-\expe{U_k}\right)^\top}\right)\\
&=\lambda_{\min}\left(H^{1/2}\expe{ \zeta_k \zeta_k^\top}H^{1/2}\right).
\end{align*}
Thus, using the fact that $|\mathcal{N}_t^{\mathsf{SE}}|=N_t$, we get
\begin{align*}
V_t&\succeq  \rho^2 \sigma^2 N_t   I + \sum_{k\in \mathcal{N}_t^{\mathsf{SE}}} W_k.
\end{align*}
Using Weyl's inequality, it immediately follows that
\begin{align}
\lambda_{\min}(V_t) \geq    \rho^2 \sigma^2 N_t- \lambda_{\max}\left(\sum_{k\in \mathcal{N}_t^{\mathsf{SE}}} W_k\right).
\end{align}
We rely on the Matrix Azuma Inequality \eqref{eq:azuma} to establish an  upper bound on $ \lambda_{\max} (\sum_{k\in \mathcal{N}_t} W_k)$, which holds with high probability. 
\begin{theore}[Matrix Azuma Inequality]{\cite[Theorem 7.1. and Remark 7.8.]{tropp2012user}}\label{thm:azuma} 
Let $\{\mathcal{F}_k\}_{k=0}^\infty$ be a filtration. Consider the random process $\{Y_k\}_{k=1}^\infty$ adapted to the filtration $\{\mathcal{F}_k\}_{k=1}^\infty$. Each $Y_k$ is a self-adjoint matrix with dimension $d$ such that
\begin{align*}
\expe{Y_k\mid \mathcal{F}_{k-1}} =0\ \text{for}\ k= 1,2,3, \ldots.,
\end{align*}
and
\begin{align*}
Y_k^2\preceq A_k^2\ \text{almost surely for}\ k= 1,2,3, \ldots,
\end{align*}
where $\{A_k\}_{k=1}^\infty$ is a sequence of deterministic matrices. Moreover, the sequence $\{Y_k\}_{k=1}^\infty$ is conditionally symmetric, i.e., $Y_k\sim -Y_k$ conditional on $\mathcal{F}_{k-1}$. Then, for all $\delta\geq 0$ and $t\geq 1$, it holds that
\begin{align}
\prob{\lambda_{\max}\left( \sum_{k=1}^t Y_k\right)\geq \delta} &\leq d\cdot \exp\left(-\frac{\delta^2}{2\lnorm{\sum_{k=1}^t A_k^2}}\right)\label{eq:azuma}.
\end{align}
\end{theore}
In order to apply the Matrix Azuma Inequality \eqref{eq:azuma}, we first show that the sequence of random matrices $\{W_k\}_{k=1}^\infty$ satisfy the assumptions of Theorem \ref{thm:azuma}. From the definition of $W_k$ in Equation \eqref{eq:w_k}, it follows that $W_k=W_k^\top$ for all $k\geq 1$. Define the filtration $\mathcal{F}_k = \sigma(\xS_1,\ldots,\xS_{k+1},\zeta_1,\ldots,\zeta_k)$ for all $k\geq 1$. It immediately follows that $W_k$ is $\mathcal{F}_k$-measurable, conditionally symmetric, and $\expe{W_k\mid \mathcal{F}_{k-1}}=0$. We now construct the sequence of deterministic matrices $\{A_k\}_{k=1}^\infty$ such that it almost surely holds that $W_k^2\preceq A_k^2$. Using the fact that the trace of a matrix is equal to the sum of its eigenvalues, it almost surely holds that
\begin{align*}
\lambda_{\max}(W_k)&\leq \tr{W_k}\\
& = 2 ((1-\rho)\xS_k + \rho \bar{x})^\top(\rho H^{1/2}\zeta_k) + \rho^2 \zeta_k^\top H \zeta_k - \rho^2\tr{H^{1/2}\expe{\zeta_k\zeta_k^\top}H^{1/2}}\\
& \leq 2 ((1-\rho)\xS_k + \rho \bar{x})^\top(\rho H^{1/2}\zeta_k) + \rho^2 \lambda_{\max}(H) - \rho^2\sigma^2 d,
\end{align*} 
where the inequality follows from the fact that $\norm{\zeta_k}=1$ for all $k\geq 1$ and the definition of $\sigma^2$. Using the fact that $\norm{\xS_k}\leq L$, and the Cauchy-Schwarz inequality, it almost surely holds that
\begin{align*}
\lambda_{\max}(W_k)&\leq k_3,
\end{align*}
where $k_3$ is defined as
\begin{align*}
k_3 &=2\rho((1-\rho)L + \rho \norm{\bar{x}}) \sqrt{\lambda_{\max}(H)}  +  \rho^2 \lambda_{\max}(H) - \rho^2\sigma^2 d.
\end{align*}
Define $A_k = k_3I$ for all $k\geq 1$. Then, it almost surely holds that $W_k^2 \preceq \lambda_{\max}(W_k)^2 I\preceq A_k^2$ for all $k\geq 1$. Thus, the sequence of random matrices $\{W_k\}_{k=1}^\infty$ satisfies all the assumptions of Theorem \ref{thm:azuma}. Using the Cauchy-Schwarz inequality, we get
\begin{align*}
\lnorm{\sum_{k\in \mathcal{N}_t^{\mathsf{SE}}} A_k^2}&\leq \sum_{k\in \mathcal{N}_t^{\mathsf{SE}}}\lnorm{A_k^2}\leq N_t k_3^2.
\end{align*}
 Using the Matrix Azuma Inequality \eqref{eq:azuma}, for any $\delta\geq 0$, it holds that
\begin{align*}
\prob{\lambda_{\max}\left(\sum_{k\in \mathcal{N}_t^{\mathsf{SE}}} W_k\right)\geq \delta\ \bigg|\ N_t = n}&\leq d\cdot \exp\left(-\frac{\delta^2}{2 n k_3^2}\right).
\end{align*}
By setting $\delta = (1-\mu) \rho^2\sigma^2 N_t  $, we get
\begin{align*}
\prob{ \lambda_{\max}\left(\sum_{k\in \mathcal{N}_t^{\mathsf{SE}}} W_k\right)\geq (1-\mu) \rho^2\sigma^2 N_t \ \bigg|\ N_t = n }&\leq d\cdot \exp\left(-\frac{(1-\mu)^2 \rho^4\sigma^4 n^2}{ 2 n k_3^2}\right)\\
&\leq d\cdot \exp\left(-k_4 (1-\mu)^2  n\right),
\end{align*}
where $k_4$ is defined as
\begin{align*}
k_4 = \frac{\rho^4\sigma^4}{ 2k_3^2}.
\end{align*}
\end{appendices}

\end{document}